\documentclass[sn-nature,Numbered]{sn-jnl}

\usepackage{graphicx}%
\usepackage{multirow}%
\usepackage{amsmath,amssymb,amsfonts}%
\usepackage{amsthm}%
\usepackage{mathrsfs}%
\usepackage[title]{appendix}%
\usepackage{xcolor}%
\usepackage{textcomp}%
\usepackage{manyfoot}%
\usepackage{booktabs}%
\usepackage{algorithm}
\usepackage{algpseudocode}
\usepackage{listings}%
\usepackage{subfigure}
\usepackage{url}
\usepackage{bm}

\theoremstyle{thmstyleone}%
\theoremstyle{thmstyletwo}%

\theoremstyle{thmstylethree}%

\begin{document}

\title[dper]{Enhancing Deep Deterministic Policy Gradients on Continuous Control Tasks with Decoupled Prioritized Experience Replay}


\author*[1]{\fnm{Mehmet Efe} \sur{Lorasdagi}}\email{lorasdagi@g.ucla.edu}

\author[2]{\fnm{Dogan Can} \sur{Cicek}}\email{cicek@ee.bilkent.edu.tr}

\author[2]{\fnm{Furkan Burak} \sur{Mutlu}}\email{furkan.mutlu@ee.bilkent.edu.tr}

\author[2]{\fnm{Suleyman Serdar} \sur{Kozat}}\email{kozat@ee.bilkent.edu.tr}

\affil*[1]{\orgdiv{Electrical and Computer Engineering}, \orgname{University of California Los Angeles}, \orgaddress{\city{Los Angeles},  \state{California}, \country{United States}}}

\affil[2]{\orgdiv{Electrical and Electronics Engineering}, \orgname{Bilkent University}, \orgaddress{\city{Ankara},\country{Turkey}}}

\abstract{\textbf{Background:} Deep Deterministic Policy Gradient-based reinforcement learning algorithms utilize Actor--Critic architectures, where both networks are typically trained using identical batches of replayed transitions. However, the learning objectives and update dynamics of the Actor and Critic differ, raising concerns about whether uniform transition usage is optimal.

\textbf{Objectives:} We aim to improve the performance of deep deterministic policy gradient algorithms by decoupling the transition batches used to train the Actor and the Critic. Our goal is to design an experience replay mechanism that provides appropriate learning signals to each component by using separate, tailored batches.

\textbf{Methods:} We introduce Decoupled Prioritized Experience Replay (DPER), a novel approach that allows independent sampling of transition batches for the Actor and the Critic. DPER can be integrated into any off-policy deep reinforcement learning algorithm that operates in continuous control domains. We combine DPER with the state-of-the-art Twin Delayed DDPG algorithm and evaluate its performance across standard continuous control benchmarks.

\textbf{Results:} DPER outperforms conventional experience replay strategies such as vanilla experience replay and prioritized experience replay in multiple MuJoCo tasks from the OpenAI Gym suite.

\textbf{Conclusions:} Our findings show that decoupling experience replay for Actor and Critic networks can enhance training dynamics and final policy quality. DPER offers a generalizable mechanism that enhances performance for a wide class of actor--critic off-policy reinforcement learning algorithms.}




\maketitle

\section{Introduction}

\label{sec:intro}

Reinforcement learning shows remarkable improvement in numerous setups and tests when combined with deep learning \cite{2020walkthrough,2013arcade,2021madras}. Deep reinforcement learning started to become more popular after it was used to tackle human control tasks on ATARI \cite{mnih2015human} and then it has been widely discussed \cite{2024drlsurvey}, even being used in generative AI \cite{2024genai,2024hitl}, and have a potential to change the world considerably \cite{2021implications}. Its success in such tough environments has been observed \cite{2022starcraft}. Also, deep reinforcement learning approaches solve distinct continuous control tasks such as human-robot collaboration \cite{9309387}, motion planning \cite{8961997}, path planning \cite{8643443},  tele-robotics and tele-operation \cite{9699095}, vehicle decision making \cite{9647929}, and endoscopic capsule robot navigation \cite{8744622}, with remarkable success. 
 
Deep reinforcement learning algorithms solve tasks with sufficiently big state and action spaces since it allows the agent to generalize the Q-values of each state-action pairs and yield a parameterized policy. Deep reinforcement learning agents train their policy by using the observation and reward information they receive \cite{Sutton1998}. However, observation information collected by the agent during consecutive time steps is highly correlated, which violates proper neural network training. The experience replay mechanism tackles the given problem by enabling the agent to reuse its past experiences, in other words, using previous transitions that it collected during the training \cite{10.1007/BF00992699}. That solution breaks the temporal correlation between transitions that are used to update the parameters of the Actor and the Critic. Experience replay mechanism enhances the performance of the deep reinforcement learning agents, provides them a sustainable return improvement and a robust policy during the learning process \cite{9094324}.

The most primitive variant of the experience replay method, Vanilla Experience Replay, samples transition from the replay buffer uniformly \cite{10.1007/BF00992699}. This approach directly assumes that each transition stored to the replay buffer is equally beneficial for the training process of the agent. On the contrary, the works show that the strategy the agent uses to choose collected transitions heavily affects the performance of the agent \cite{fujimoto2019off,vanhasselt2018deep}. Developing better experience replay mechanisms is an active research field, and there are studies that surpass the performance of the Vanilla Experience Replay on continuous control tasks \cite{schaul2016prioritized,zha2019experience,zhang2018deeper,9643162,2023FBM}.

In this work, we propose a novel approach to prioritization of the collected transitions, Decoupled Prioritized Experience Replay, DPER. In our work, we decouple the training of the Actor and the Critic of the Deep Deterministic Policy Gradient Algorithms in terms of the batches of transitions that they use during the learning process. Our algorithm is usable for every off-policy deep deterministic reinforcement learning algorithm that works on continuous control tasks. We aim to show that it is possible to improve the performance of a deep reinforcement learning algorithm that has two cascaded neural structures, the Actor and the Critic, by updating them with different batches of transitions. To support our aim, we use two different Experience Replay methods in terms of their prioritization strategy. We show that a combination of two different Experience Replay algorithms can yield better results in particular continuous control tasks.

We evaluate our algorithm, DPER, by coupling it with the TD3 algorithm \cite{fujimoto2018addressing}. We collate DPER with Vanilla Experience Replay and Prioritized Experience Replay. We use OpenAI gym and MuJoCo learning environments, which consist of a wide range of robot control tasks to test the performance of our proposed algorithm \cite{brockman2016openai,todorov_erez_tassa_2012}.

We summarize our main contributions in this paper as follows:
\begin{itemize}
\item DPER is the first work that decouples the Actor and the Critic training by using a different batch of transitions,
\item Develop DPER to make the Critic learn through transitions that yield more temporal difference error, while the Actor uses more on-policy batches of transition to update its parameters.
\item We show that a combination of two experience replay methods can outperform the cases that they used separately.
\item We study and analyze the metrics used for the prioritization process of the transitions for both the Actor and the Critic training. 
\item We demonstrate that DPER brings significant performance improvements to the state-of-the-art deep deterministic policy gradients algorithm TD3 on continuous control tasks.
\end{itemize}

\section{Background}
\label{sec:background}
In this section, we give a brief introduction to reinforcement learning and summarize two deep deterministic policy gradient algorithms designed for solving continuous control tasks. We also briefly cover the current experience replay methods in the literature.

\subsection{Reinforcement Learning}

A reinforcement learning agent optimizes its policy to obtain the maximum amount of return by consecutive trial and error. Reinforcement Learning tasks are defined as Markov Decision Processes (MDP) \cite{Sutton1998}. During the learning process, the agent receives state information $s \in \mathcal{S}$. After observing the state information, the agent chooses its action from the action space $\mathcal{A}$, $a_t \in \mathcal{A}$, with respect to its policy $a_t \sim \pi(a|s_t)$. Then, the agent applies the selected action to the environment and receives the reward, $r$, and the next state information, $s' \in \mathcal{S}$ from the environment. The environment gives the reward and the next state to the agent according to the probability distribution $P(s', r|s, a)$. 

The collected information throughout the above cycle is defined as a transition, $(s, a, r, s')$. Agents that work with experience replay mechanisms keep these transitions in their replay memory and use them to optimize the return function:
\begin{equation}
    R_{t} = E_{\pi}[\sum_{i = t}^{T}\gamma^{i - t}r(s_{i}, a_{i})],
\end{equation}
 where $\gamma$ represents the discount factor.

\subsection{Deep Deterministic Policy Gradient}

One of the first deterministic deep reinforcement learning algorithms that tackle tasks with continuous action space is the Deep Deterministic Policy Gradients (DDPG) \cite{lillicrap2019continuous}. The DDPG algorithm has two cascaded deep neural network structures named the Critic and the Actor. The Critic estimates the Q-value of the given state-action pair as follows: 
\begin{equation}
    y = C(s, a| v),
\end{equation} where $C$ is the Critic, $v$ is the parameters of the Critic, and $y$ is the Q-value of the given state-action pairs. The property of the Critic is that estimating Q-value by taking state and action information as the input enables the algorithm to work correctly on continuous action spaces. On the other hand, the Actor controls the agent's behavior with a parametric policy. So, it determines which action should be taken by the agent given the state information:
\begin{equation}
    a = A(s| w),
\end{equation} where $A$ is the Actor, $w$ is the parameters of the Actor, and $a$ is the action that is produced by the Actor. In addition to these aforementioned networks, the DDPG algorithm also has one more nested neural network structure named Target Network. This nested neural structure includes the Actor Target, which shares the same weights with the Actor at the beginning of the training. Also, the Critic Target has the same weights as the Critic at the start of the training. The Critic and the Actor updates are consecutive for deep deterministic policy gradients algorithms. In the conventional update method for these networks, first, a one step temporal difference error is calculated, and the loss function is defined with respect to this value:
\begin{equation}
    \mathcal{L}(v) = \mathbb{E}_{(s, a, r, s') \sim \mathcal{B}}[\mathcal{Y}- C(s, a| v)]^{2},
\end{equation} where $\mathcal{B}$ is the experiences that the agent collected, the target value for Critic is defined as:
\begin{equation}
    \mathcal{Y} = r + \gamma C'(s', A(s'|w')| v'),
\end{equation}
where $C'$ is the Critic Target and $w'$ and $v'$ are the weights of the actor and critic target networks, respectively. The output of the Critic can be assumed as the objective function of the agent, and it represents the discounted cumulative reward of the agent at a given state by following the policy that is parameterized by the Actor:
\begin{equation}
    J(v,w) = \mathbb{E}_{s \sim \mathcal{B}}[C(s, a| v)|_{a = A(s| w)}A(s| w)],
\end{equation}
Therefore, by taking the gradient of the $J(v,w)$ with respect to the Actor's weights, the Actor is updated:
\begin{equation}
    \nabla_{w}J(v,w) = \mathbb{E}_{s \sim \mathcal{B}}[\nabla_{a}C(s, a| v)|_{a = A(s| w)}\nabla_{w}A(s| w)],
\end{equation}
It has been shown that bootstrapping from the same neural network significantly harms the training process \cite{vanHasselt_Guez_Silver_2016}. To remedy the given issue, the DDPG algorithm utilizes Target-Networks. The Actor Target and the Critic Target are softly updated by using the Actor and the Critic:
\begin{equation}
    w' = \tau w + (1 - \tau)w', \quad v' = \tau v + (1 - \tau)v',
\end{equation}
where $\tau$ is the rate of the soft update.
\subsection{Twin Delayed DDPG}
Studies suggest that even though the DDPG algorithm performs well on continuous control tasks, it suffers from an overestimation problem \cite{fujimoto2018addressing,lillicrap2019continuous}. Overestimation problem is defined as estimating the Q-value of a given state-action pair, $(s,a)$, while following a policy, $w$, higher than it should be:
\begin{equation}
    C_\pi(s,a|v) > Q_\pi(s,a),
\end{equation} where $C_\pi$ is the estimated Q-value of the given state-action pair while following the policy $\pi$, and $Q_\pi$ is the true Q-value of the given state-action pair while following the policy $\pi$. Despite the presence of the Target Networks, in some cases, bootstrapping on overestimated state-action pairs dominantly accelerates the overestimation problem and cripples the learning process \cite{2022Icarte}. Moreover, it may lead to the agent instantly forgetting its policy in the middle of the training \cite{vanhasselt2018deep}. TD3 remedies this problem by adding one more Critic to the neural network structure \cite{fujimoto2018addressing}. TD3 algorithm formulates the target value for the Q-value of a given state-action pair as follows:
\begin{equation}
    \mathcal{Y} = r + \gamma \underset{i=1, 2}{\mathrm{min}}C'_{i}(s', A'(s'| w')| v'_{i}),
\end{equation}
and the authors call this procedure Clipped Double Q-Learning \cite{fujimoto2018addressing}. This advancement enables the TD3 to dominantly outperform the DDPG in continuous control tasks. The TD3 solves particular tasks that the DDPG cannot find any decent policy.

{Moreover, TD3 incorporates exploration noise $\epsilon$ to the predicted action $A(s|w)$ to improve the exploration of the agent in the state-action space \cite{fujimoto2018addressing}. This is common practice in many off-policy algorithms \cite{lillicrap2019continuous}.
Thus, the realized action at each step becomes $A(s|w) + \epsilon$, where $\epsilon \sim N(0, \sigma)$. Hence, the realized action is a random variable, distributed with $N(A(s|w), \sigma)$.}

\subsection{Experience Replay Mechanism}
Neural networks yield much better results when the training dataset includes mostly uncorrelated samples. However, in reinforcement learning, the agent receives state, and next state information is overlapped for successive transitions, i.e., the agent's next state, $s'$, at time step $t$, is the same as the state, $s$, at the time step $t+1$. Then, training deep reinforcement learning is challenging since the agent updates its policy and value function using heavily correlated samples from the environment. 

The experience replay mechanism remedies the given fact by adding a cyclic replay buffer to the algorithms. The first experience replay method stores the experiences that the agent collects while exploring the environment to the replay buffer as transitions. Then, the agent updates its parameters using the samples selected from the replay buffer to break the correlation between samples that are used for the training at a time. The most primitive method, Vanilla Experience Replay, samples transitions from the replay uniformly.

Prioritized Experience Replay is one of the most well-known techniques used for assigning a different level of importance for the transitions \cite{schaul2016prioritized}. PER increases the tendency to feed the value function of the deep reinforcement learning algorithm for the transitions collected while the agent encounters unexpected outcomes. PER measures the unexpectedness level of a transition by using temporal difference error as a proxy. 

Transitions in the replay buffer can be sampled by following heuristic rules. Focused Experience Replay introduces an alternative probability distribution, half normal distribution, to increase the sampling probabilities of the recently added transitions to the replay buffer \cite{9444458}. Hindsight Experience Replay, HER, is an experience reuse mechanism that provides sub-goals for the agent to divide the main task into smaller tasks through experience replay \cite{2017hindsight,2023Xu}. 

Experience Replay can also be formulated as a learning problem that proceeds parallel with the training of the deep reinforcement learning agent \cite{zha2019experience}. By utilizing a parameterized experience reuse strategy, the Neural Experience Replay Sampler technique manages experience replay prioritization \cite{oh2021learning}. This method gives scores that determine the importance level of the transitions when the corresponding transition is collected with respect to the replay policy.

\section{Motivation}
In this section, we examine the limitations that may arise when using prioritized experience replay, particularly as proposed in the PER algorithm. We also outline the rationale for decoupling the training of actor and critic networks by utilizing different transition batches during their updates. Finally, we discuss the potential benefits of minimizing the off-policyness in reinforcement learning algorithms.

\subsection{Advantages and Disadvantages of Prioritizing Experience Replay}

In actor-critic models, there is usually a sequential update mechanism for the networks: the critic is optimized first to produce accurate Q-value predictions, then the actor updates its policy parameters using feedback from the critic, i.e., the value predictions of critic are used to estimate the policy gradient. When Prioritized Experience Replay is used, each transition’s sampling probability $(p_i)$ is set proportional to the absolute value of its TD error $|(\delta_i)|$, as shown below:
\begin{equation}
        \delta_i = r(s_i,a_i,s'_i) + \gamma C'\Bigl(s'_i, A'(s'_i\mid w') \mid v'\Bigr) - C(s_i, a_i \mid v), \quad p_i \propto |\delta_i|.
\end{equation}

Focusing on transitions with higher TD errors accelerates the critic’s learning as higher TD error directly indicates that the critic estimates the value of the specific state poorly. However, it can also mislead the actor’s training. When the sampled batches are dominated by such transitions, the actor may end up learning from unreliable value estimates provided by the critic, ultimately degrading the policy learning. Thus, while PER supports faster convergence of the critic, it might adversely affect the actor's optimization trajectory.

\subsection{Disadvantages of the Off-Policy Learning}
In a standard reinforcement learning setting, an agent must explore a wide array of state-action pairs to derive a satisfactory policy for the given environment. However, this exploration becomes impractical in environments with large state and action spaces. Here, neural networks act as powerful approximators, allowing the agent to generalize Q-values across unseen state-action pairs, which becomes essential in managing such complex problems.

Vanilla Experience Replay, the most basic form of experience replay, assigns equal priority to each transition, which implies it contains the assumption that all stored events have similar importance for the training of the agent. Hence, it may seem straightforward to design a prioritization scheme with higher performance. Nevertheless, studies have shown that altering the replay distribution can increase the off-policyness of the algorithm, potentially degrading its performance \cite{vanhasselt2018deep,zhang2018deeper}. Research examining these modifications has found that even buffer size in cyclic replay mechanisms significantly affects learning outcomes, since it impacts the degree of policy divergence over time \cite{zhang2018deeper}. One can lower the off-policy characteristics of learning by suitably changing transition sampling probabilities, improving performance on several tasks \cite{vanhasselt2018deep}. Limiting off-policyness is essential since higher degrees of policy divergence could cause unstable updates, compromising learning stability.

This challenge is even more pronounced in actor-critic systems. Although critic updates benefit from a diverse range of transitions, including those with large TD errors, the actor is especially sensitive to mismatches between stored actions (based on older policies) and the actions suggested by the present policy. Such misalignments can cause false gradient estimates, which then misdirect policy changes and lower general performance \cite{2023FBM}. Unlike Q-learning methods that directly use value estimates for action selection and have no separate policy component, actor-critic models involve an explicit actor network that can suffer from such discrepancies. In fact, off-policyness can even be advantageous in Q-learning-based systems, particularly in relation to on-policy approaches such as SARSA \cite{Sutton1998}. Excessive off-policyness, however, introduces instability during the actor's training phase in Deep Deterministic Policy Gradient and other actor-critic approaches. Both theoretical and empirical studies \cite{2023FBM} underline the need of maintaining the behavior policy, shown by the sampled transitions, in close alignment with the current policy to ensure consistent and stable updates for the actor.

\section{Decoupled Prioritized Experience Replay}
\label{sec:dper}
In this section, we present the architecture of our proposed algorithm, DPER, and further elaborate on how it selects distinct transition batches for training the actor and critic networks.

\subsection{Reducing the Off-policyness of the Learning Algorithm}
Each transition stored in the experience replay buffer originates from the behavior policy the agent follows at the time of data collection, denoted by $A_t(\cdot)$. Although it is theoretically possible to retain the entire actor model that generated each individual transition, this is practically infeasible due to the large number of parameters (often in the millions) and the scale of transitions stored. To address this, our algorithm treats each sampled batch of transitions as if it were generated by a stochastic policy. This treatment is based on the fact that executed actions are determined by adding exploratory noise to the prediction of the actor, i.e., indeed we have a stochastic policy. We define the most probable policy responsible for generating a given batch as the Transition Generator, denoted by $\lambda$. To estimate this policy, we utilize the policy of the agent at the corresponding time step. Our method forwards the batch of states from the replay buffer through the current actor network to obtain the actions the agent would take in those states at time $t$:
\begin{equation}
    \boldsymbol{\hat{X}}^{b\times m} = A_t(\boldsymbol{S}^{b \times n}| w),
\end{equation}
Here, $b$ is the batch size, $A_t(\cdot|w)$ is the actor parameterized by weights $w$, $m$ is the dimensionality of the action space, and $n$ is the dimensionality of the state space. We then compute the difference between these current actions and the actions stored with the transitions in the buffer to approximate the deviation from the original behavior policy:
\begin{equation}
    \boldsymbol{\dot{X}}^{b\times m} := \boldsymbol{\hat{X}}^{b\times m} - \boldsymbol{X}^{b\times m},
\end{equation}
where $\boldsymbol{X}^{b\times m}$ contains the action components from the stored transitions. As directly recording the exact Transition Generator would require storing all actor weights at each timestep, an impractical approach, prior work \cite{9643162} suggests approximating this generator as a stochastic policy modeled as a probability distribution over the action space.

We determine the mean of this distribution by calculating the average deviation between stored actions and those predicted by the current actor for the same states:
\begin{equation}
    \mu_{\lambda}^{1 \times m} = \frac{1}{b} \sum_{i \in b} \boldsymbol{\dot{X}}^{b\times m}_{ij},
\end{equation}
where $i$ indexes transitions and $j$ indexes action dimensions. Based on this, we define the covariance matrix of the Transition Generator as:
\begin{equation}
    \mathrm{\Large\Sigma}_{\lambda}^{m \times m} = \frac{1}{b-1} \sum_{k \in b}  (\boldsymbol{\dot{x}}_k^{1 \times m}-\mu_{\lambda}^{1 \times m})^{\top}(\boldsymbol{\dot{x}}_k^{1 \times m}-\mu_{\lambda}^{1 \times m}),
\end{equation}
We represent this Transition Generator using a multivariate normal distribution to ensure maximal entropy:
\begin{equation}
    \lambda \sim N(\mu_{\lambda}^{1 \times m}, \mathrm{\Large\Sigma}_{\lambda}^{m \times m}).
\end{equation}

To evaluate the similarity between the agent’s current behavior policy and the batch’s original generating policy, we use the Kullback–Leibler (KL) divergence. The KL score for each batch is computed as:
\begin{equation}
    \eta = \mathrm{D}_{\mathrm{KL}}(N(\mu_\lambda, \Sigma_\lambda) \| N(0, \sigma \mathbb{I})),
\end{equation}
Here, $\mathbb{I}$ denotes the identity matrix, and $N(0, \sigma \mathbb{I})$ captures the agent’s exploration noise. We interpret a Transition Generator with zero mean and diagonal covariance matrix matching the exploration noise as representing a perfectly on-policy batch.

This formulation also aligns with minimizing the mean squared error between the stored actions and those predicted by the actor. Recall our representation of the Transition Generator:
\begin{equation}
{\mu_\lambda \;=\;\frac{1}{b}\sum_{i=1}^b\!\bigl(A_t(s_i\mid w)-a_i\bigr),}
\end{equation}
\begin{equation}
{\Sigma_\lambda \;=\;\frac{1}{b-1}\sum_{i=1}^b\bigl(A_t(s_i\mid w)-a_i-\mu_\lambda\bigr)^\top \bigl(A_t(s_i\mid w)-a_i-\mu_\lambda\bigr).}
\end{equation}
We again use the KL divergence to measure deviation from the exploration policy \(\mathcal{N}(0,\sigma \mathbb{I})\):
This expands to:
\begin{equation}
\begin{aligned}
{
\eta}
&=
{\frac{1}{2}\Bigl[
\mathrm{tr}\bigl(\sigma^{-1}\,\Sigma_\lambda\bigr)
+\,\sigma^{-1}\,\mu_\lambda^\top\mu_\lambda
-\,m
+\ln\frac{\det(\sigma  \mathbb{I})}{\det(\Sigma_\lambda)}
\Bigr]}
\\
&=
{\underbrace{\frac{1}{2\sigma}\,\mu_\lambda^\top\mu_\lambda}_{\displaystyle\text{(A)}}
\;+\;
\underbrace{\frac{1}{2}\Bigl[\mathrm{tr}(\sigma^{-1}\Sigma_\lambda)-m
+\ln\frac{\sigma^{m}}{\det(\Sigma_\lambda)}\Bigr]}_{\displaystyle\text{(B)}}.}
\end{aligned}
\end{equation}

\textbf{Equivalence under Diagonal Covariance Assumption.}
Assuming \(\Sigma_\lambda \approx \sigma \mathbb{I}\), justified if action noise is identically and independently distributed with scalar variance $\sigma$, the term (B) vanishes because \(\mathrm{tr}( \mathbb{I})=m\) and \(\det(\sigma \mathbb{I})=\sigma^{m}\). This yields:
\begin{equation}
\eta
\;=\;
\frac{1}{2\sigma}\,\|\mu_\lambda\|_2^2.
\end{equation}
Given \(\mu_\lambda=\tfrac1b\sum_{i=1}^b\bigl(A_t(s_i\mid w)-a_i\bigr),\) it follows that:
\begin{equation}
\|\mu_\lambda\|_2^2
=\Bigl\|\frac{1}{b}\sum_{i=1}^b\!\bigl(A_t(s_i\mid w)-a_i\bigr)\Bigr\|_2^2
\approx
\frac{1}{b}\sum_{i=1}^b\|A_t(s_i\mid w)-a_i\|_2^2.
\end{equation}
Thus, minimizing \(\eta\) approximately corresponds to selecting the batch with the lowest mean squared error in action prediction under this assumption.

However, identifying a batch that strictly satisfies this optimality condition is non-trivial. Consequently, we define a maximum number of sampling trials $K$, and among the $K$ candidate batches, our algorithm selects the one minimizing $\eta$. This ensures that the actor is updated using batches that closely resemble on-policy behavior.

\subsection{Training the Actor and the Critic with Different Batch of Transitions }
For updating the critic network, our algorithm incorporates the proportional variant of Prioritized Experience Replay (PER) \cite{schaul2016prioritized}. This method assigns higher sampling probabilities to transitions with greater temporal difference (TD) errors:
\begin{equation}
    \delta =|r(s,a,s')+\gamma C'\left(s^{\prime}, a^{\prime} | v^{\prime}\right)-C(s, a | v)|,
\end{equation}
where $\delta$ denotes the TD error. In DPER, each transition is assigned a priority immediately upon collection. Continuously updating all priorities after each timestep is computationally expensive, especially when the buffer size is large. Therefore, like in PER, we maintain initial priorities until the transitions are sampled again.

Initially, this procedure can face a challenge: a transition stored with a small temporal difference error may be sampled infrequently, or not at all, even though its significance can increase as the critic and critic target parameters change during training. To prevent this oversight, our algorithm utilizes a bias correction parameter, $\alpha$, to prioritize transitions for critic training, adopting the strategy from the proportional Prioritized Experience Replay method.

\begin{equation}
P(i)=\frac{\delta_{i}^{\alpha}}{\sum_{k} \delta_{k}^{\alpha}},
\end{equation}
where $\delta_i$ and $\delta_k$ are the TD errors for transitions $i$ and $k$, respectively. This approach allows us to preserve PER’s advantages while ensuring DPER remains scalable and effective for training the critic network.

We summarize our algorithm in Algorithm \ref{alg:DPER}.

\begin{algorithm}[htbp] \label{alg:DPER}
    \caption{DPER}
    \begin{algorithmic}[1]
        \State \textbf{Input:} Batch size $b$, policy delay $M$, maximum candidate batches $K$, actor $A_t(\cdot|w)$, exploration noise covariance $\sigma$, and other standard RL hyperparameters
        \State Initialize replay buffer $\mathcal{B}$
        \For{$t = 1,2,\dots,T$}
            \State Observe state $s_t$, select action $a_t \sim \pi_{w}(a|s_t) + \epsilon$
            \State Observe reward $r_t$ and next state $s'_t$
            \State Store $(s_t,a_t,r_t,s'_t)$ in $\mathcal{B}$
            \State Sample mini-batch $D$ from $\mathcal{B}$ and update the critic using standard TD-error methods
            \If{$t \bmod M = 0$}
                \For{$n=1$ \textbf{to} $K$}
                    \State Sample candidate mini-batch $D_n$ of size $b$ from $\mathcal{B}$
                    \State Compute current actions for states in $D_n$: $\hat{\bm{X}}_n = A_t(\bm{S}_n|w)$
                    \State Retrieve stored actions $\bm{X}_n$ from $D_n$ and compute $\dot{\bm{X}}_n = \hat{\bm{X}}_n - \bm{X}_n$
                    \State Compute $\mu_{\lambda,n}$ and $\Sigma_{\lambda,n}$ from $\dot{\bm{X}}_n$
                    \State Compute KL divergence $\eta_n$
                \EndFor
                \State $n^* = \arg\min_{1\le n\le K} \eta_n$ and set $D_{n^*}$ as the batch for actor update
                \State Update the actor and target networks using $D_{n^*}$
            \EndIf
        \EndFor
    \end{algorithmic}
\end{algorithm}

\section{Implementation Details}
\label{sec:implementation}
We assess our introduced algorithm against PER and the vanilla Experience Replay. We run ten independent runs with the same random seeds for every method to guarantee a fair comparison. The PER algorithm is implemented following the official reference implementations. Specifically, we utilize the proportional variant of Prioritized Experience Replay for evaluating our approach, as introduced by \cite{schaul2016prioritized}.

To benchmark DPER, we integrate it, as well as the other replay strategies used for comparative evaluation, within the TD3 framework. We adopt the original implementation of TD3 provided in its foundational work \cite{fujimoto2018addressing}. In this setup, both the actor and critic networks consist of three layers, with the two hidden layers each containing 256 units.

The Adam optimizer \cite{kingma2017adam} is employed for updating both the actor and the critic networks. TD3 may prematurely converge to suboptimal policies in the absence of sufficient exploration. Therefore, for each experience replay strategy under evaluation, we pre-fill the replay buffer with 25,000 transitions, collected as the agent explores the environment randomly before learning begins.

Since TD3 incorporates target policy smoothing via noise injection during the critic’s target value computation, we use a noise scale of $\sigma = 0.2$ for this purpose.
DPER and the alternative replay strategies are evaluated across six continuous control environments from OpenAI Gym and MuJoCo. These tasks were selected to represent diversity in terms of both state and action space dimensions. Evaluation occurs every 1,000 timesteps, where performance is measured using the cumulative reward metric. At each evaluation interval, agents deploy their current policy for ten separate episodes. The return for each agent is computed as the average cumulative reward across these ten evaluation episodes.

\section{Experimental Results}
\label{sec:results}

In this section, we evaluate the empirical performance of our proposed DPER algorithm. We conduct a comparative analysis against two standard baselines: vanilla Experience Replay (ER) and Prioritized Experience Replay (PER). Furthermore, we perform an ablation study on the hyperparameter $K$, the number of candidate batches sampled for the actor update, to assess its impact on learning performance. All replay strategies are integrated into the TD3 framework , and experiments are conducted across six continuous control environments to ensure a comprehensive evaluation. For each configuration, we report the average cumulative reward over ten independent runs with different random seeds, smoothing the results with a window of 100 steps for improved clarity.

\subsection{Comparative Performance Analysis}

\begin{figure}[t]
  \centering

  \subfigure[Ant-v2]{%
    \includegraphics[width=0.32\textwidth]{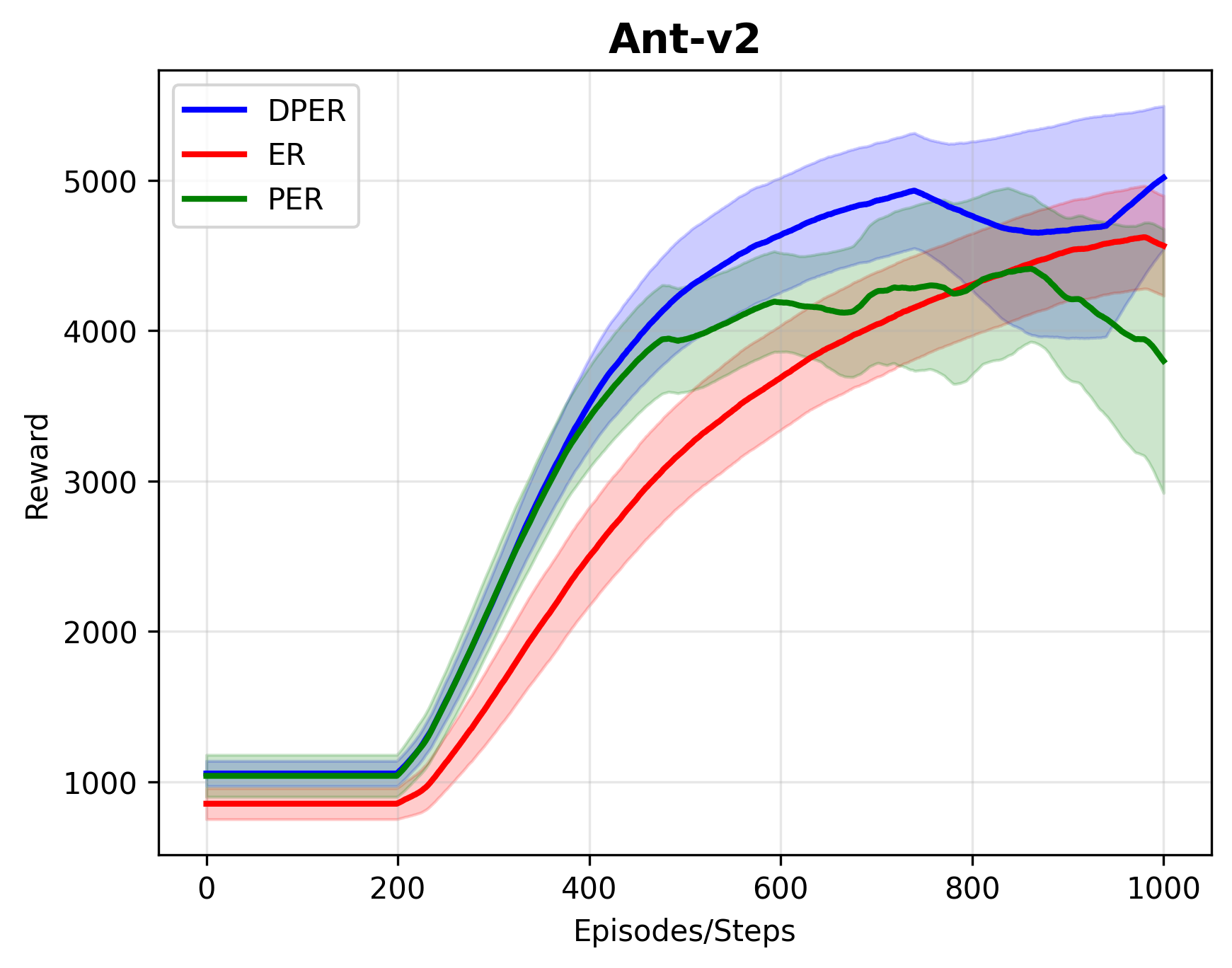}%
    \label{fig:reward_ant}
  }\hfill
  \subfigure[BipedalWalker-v3]{%
    \includegraphics[width=0.32\textwidth]{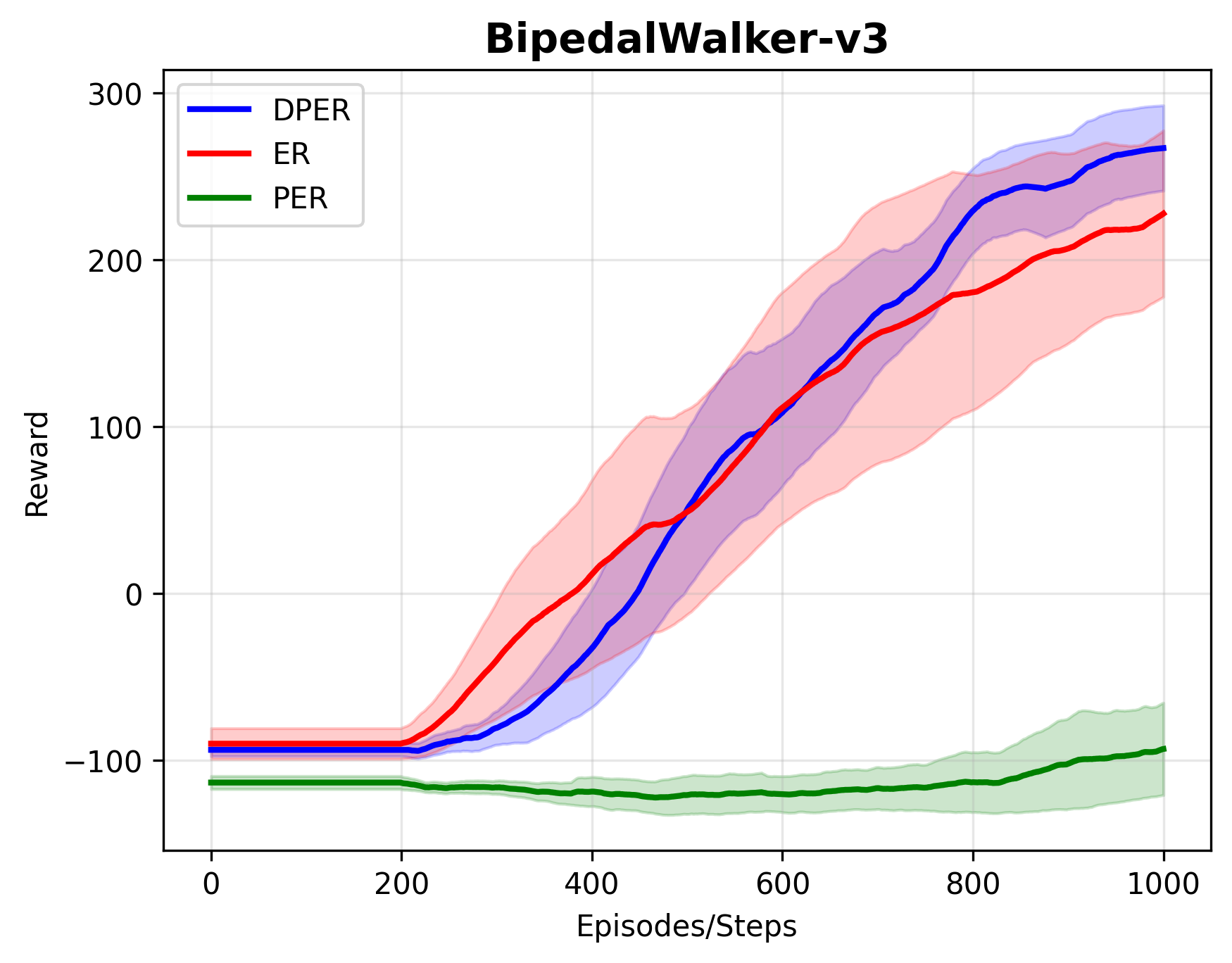}%
    \label{fig:reward_bw}
  }\hfill
  \subfigure[HalfCheetah-v2]{%
    \includegraphics[width=0.32\textwidth]{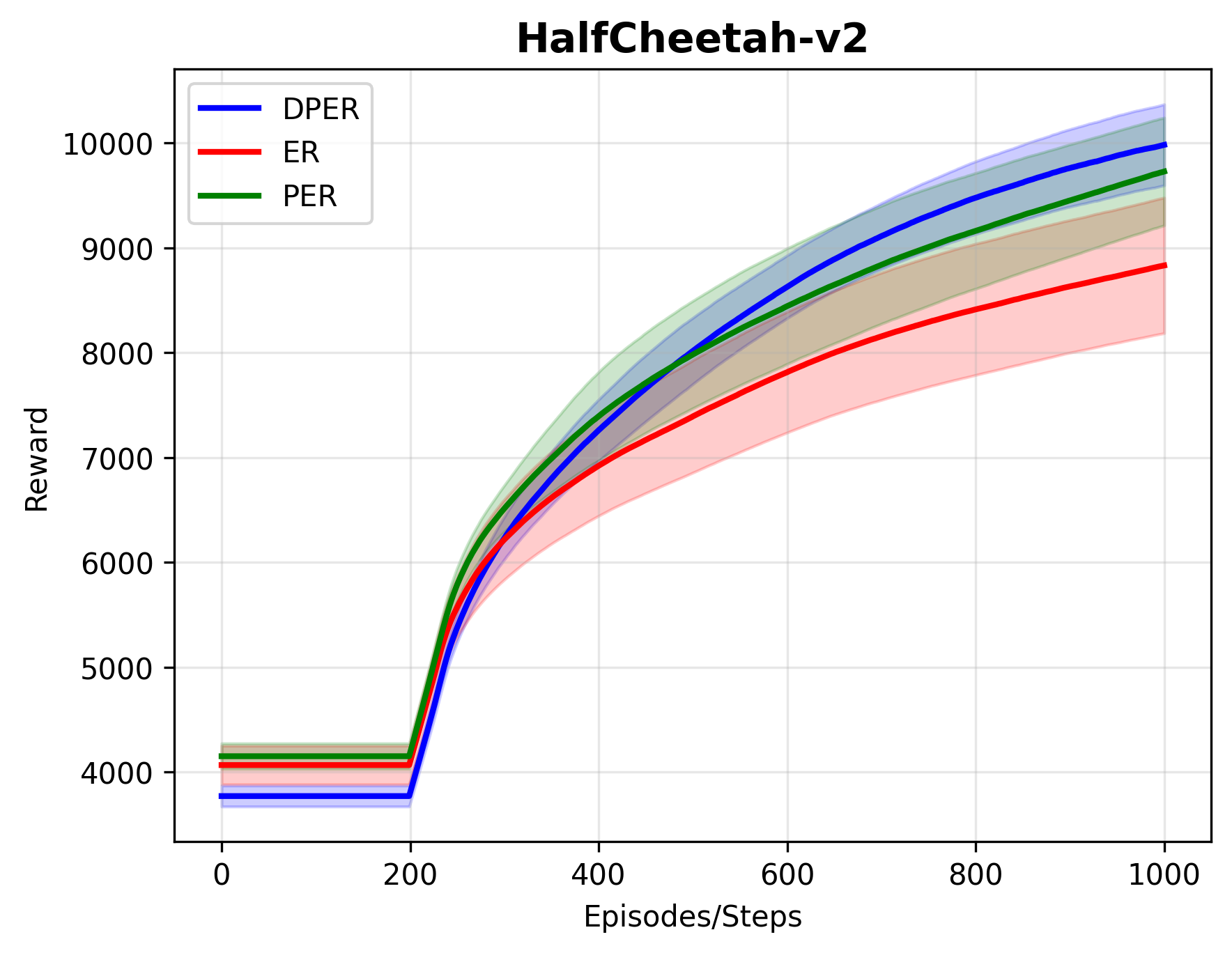}%
    \label{fig:reward_hc}
  }

  \vspace{0.6em}

  \subfigure[Hopper-v2]{%
    \includegraphics[width=0.32\textwidth]{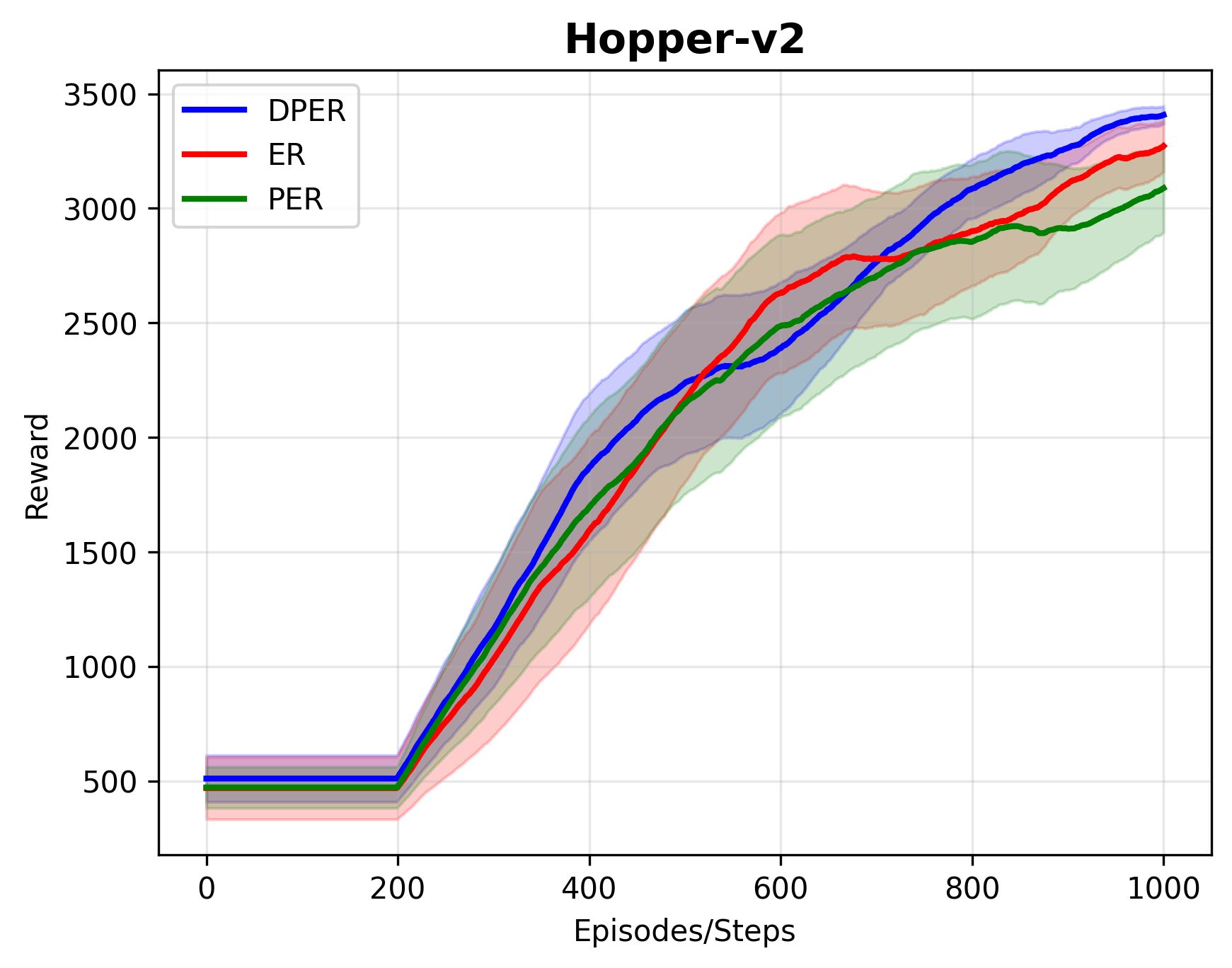}%
    \label{fig:reward_hopper}
  }\hfill
  \subfigure[LunarLanderContinuous-v2]{%
    \includegraphics[width=0.32\textwidth]{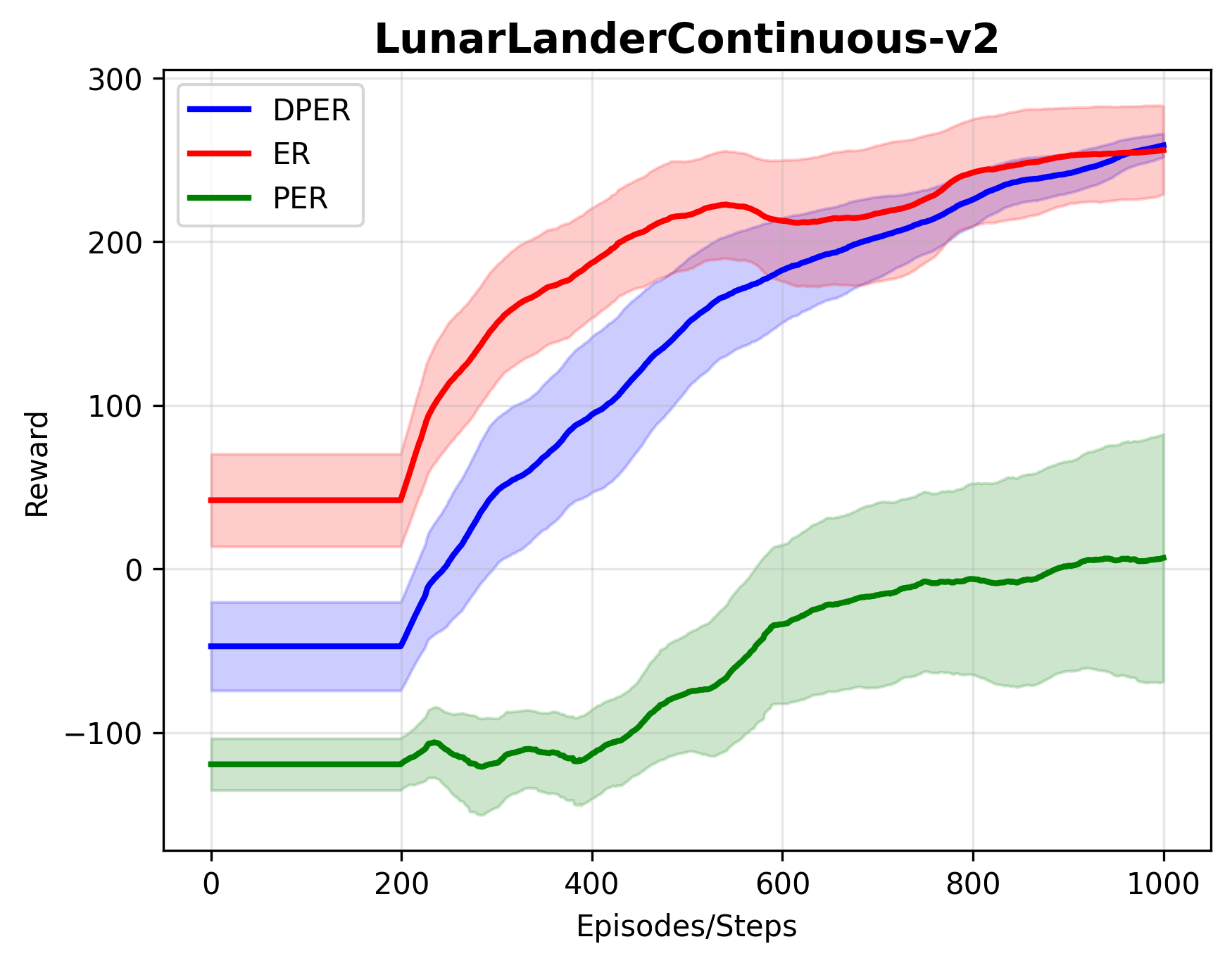}%
    \label{fig:reward_llc}
  }\hfill
  \subfigure[Walker2d-v2]{%
    \includegraphics[width=0.32\textwidth]{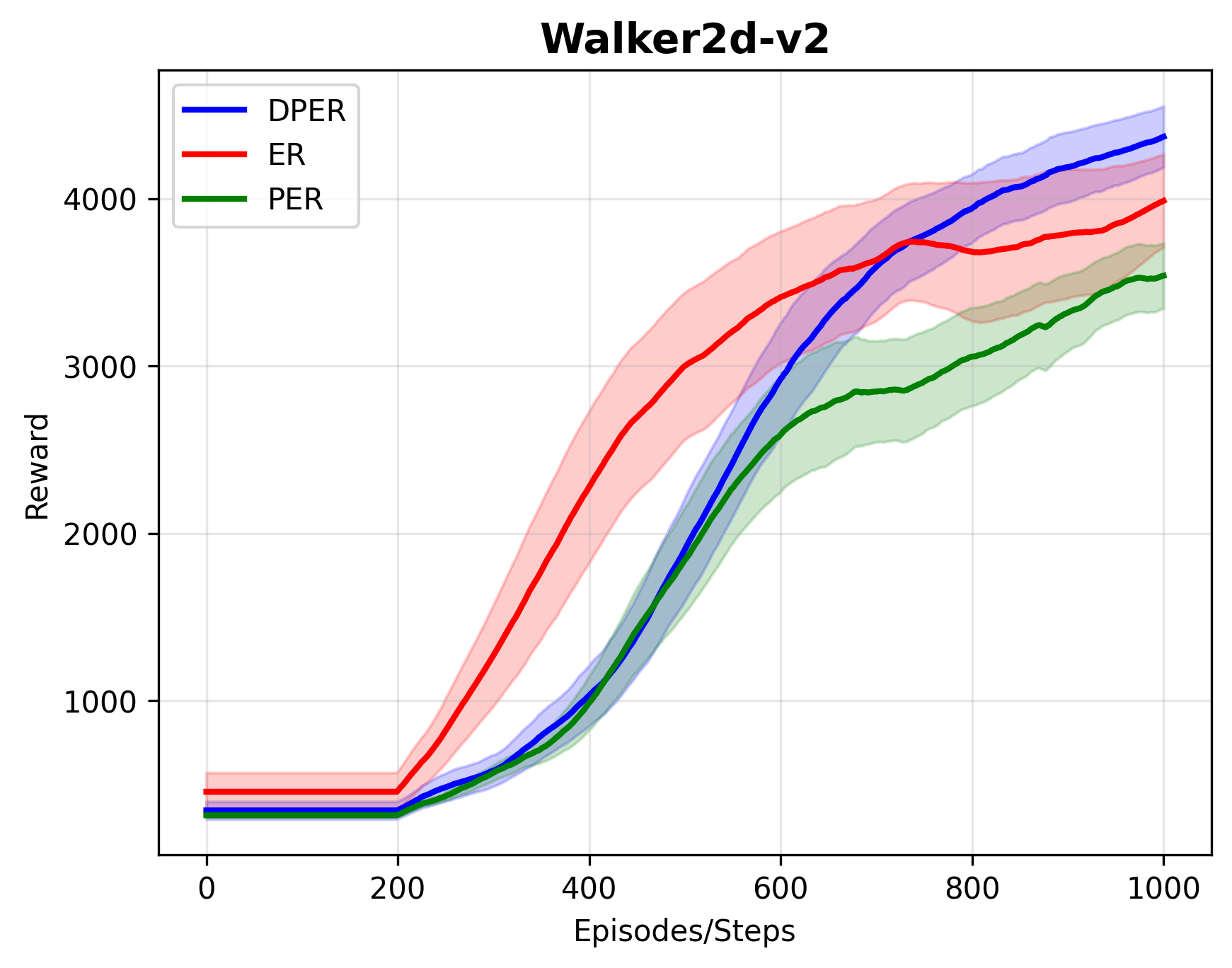}%
    \label{fig:reward_walker}
  }

  \caption{Average cumulative rewards across tasks; shaded regions indicate half standard deviation.}
  \label{fig:rewards_all}
\end{figure}

The primary performance comparison across all six environments is presented in Fig. \ref{fig:rewards_all}. For DPER, we plot the results from the best-performing $K$ value in each respective task. The results demonstrate a clear and consistent performance advantage for DPER over both ER and PER in the majority of the environments. This supports our central insight: decoupling the sampling strategies for the actor and critic networks enhances learning stability and overall performance in complex, high-dimensional control tasks. By selecting batches for the actor that more closely resemble the current policy, DPER mitigates the risks of unstable updates arising from significant policy divergence, a known challenge in off-policy actor-critic algorithms.

Notably, the performance of PER is highly inconsistent. While it surpasses ER in {Ant-v2} and {HalfCheetah-v2} , it exhibits significant performance degradation in {BipedalWalker-v3} and {LunarLanderContinuous-v2}, failing to converge to a meaningful policy. This outcome provides strong empirical evidence for the theoretical drawback we identified: prioritizing transitions solely based on high TD error can destabilize learning for actor. In environments requiring precise balance and control, such as {BipedalWalker-v3}, the critic's initial value estimates for novel or difficult states can be unreliable. By forcing the actor to learn from these transitions, PER can be misled by inaccurate gradient estimates, thus destabilizing the learning process. DPER avoids this pitfall, leading to superior performance. In the {LunarLanderContinuous-v2} task, DPER performs on par with vanilla ER. This suggests that for environments with lower dimensional state-action spaces and less complex dynamics, the destabilizing effect of off-policy updates is less pronounced, diminishing the relative advantage of DPER's more conservative actor update.

\subsection{Sensitivity to $K$}
We varied the number of candidate Actor batches $K\in\{2,3,4,5\}$.
Across tasks we do not observe a consistent monotonic gain from larger $K$; importantly,
small $K$ (2--3) already captures most of the benefit of on-policy batch selection.
Given the extra sampling overhead, $K\in\{2,3\}$ offers the best trade-off.
Detailed per-environment results are deferred to \ref{app:k} for readability.

\subsection{Critic Sampling: PER vs.\ uniform}
Swapping the Critic's sampler between PER and uniform leaves the overall picture intact: both variants learn reliably, with task-dependent wins (e.g., uniform stronger in Hopper and Walker2d, PER stronger in Ant). This indicates that the salient gain comes from \emph{decoupling} the Actor's batch selection from the Critic's sampler,
rather than from a specific choice of Critic sampling. Full numerical results are reported in \ref{app:critic-sampler}.

\subsection{Runtime}
Over $1\text{M}$ environment steps, DPER incurs a modest wall-clock overhead relative to ER and PER. The dominant cost remains forward/backward passes; drawing $K{+}1$ candidate batches for the Actor contributes little.
Thus, DPER improves stability with minimal runtime penalty, especially for small $K$.
A detailed runtime breakdown is provided in \ref{app:runtime}.

\section{Conclusion}
\label{sec:conclusion}
We propose a method that addresses the drawbacks of other conventional experience replay mechanisms. We improve two mutually exclusive experience replay methods by using them on the Actor and the Critic training separately. We introduce DPER, which decouples the training principles of the Actor and the Critic. Results show that DPER presents promising improvements and outperforms PER, and Vanilla ER on particular continuous control tasks. Also, we state that the average magnitude of the temporal difference error and KL scores yielded by the batch of transitions that are used during the training has crucial importance for the learning process.

\backmatter

\bmhead{Acknowledgements}

This study is supported by Turk Telekom within the framework of 5G and Beyond Joint Graduate Support Programme coordinated by the Information and Communication Technologies Authority.
This work is supported by The Scientific and Technological Research Council of Türkiye (TÜBİTAK) 1515 Frontier R\&D Laboratories Support Program for Turk Telekom neXt Generation Technologies Lab (XGeNTT) under project number 5249902.

\section*{Declarations}

The authors have no competing interest or conflict of interest.


\begin{appendices}

\section{Sensitivity to $K$}
\label{app:k}

\begin{figure}[t]
  \centering

  \subfigure[Ant-v2]{%
    \includegraphics[width=0.32\textwidth]{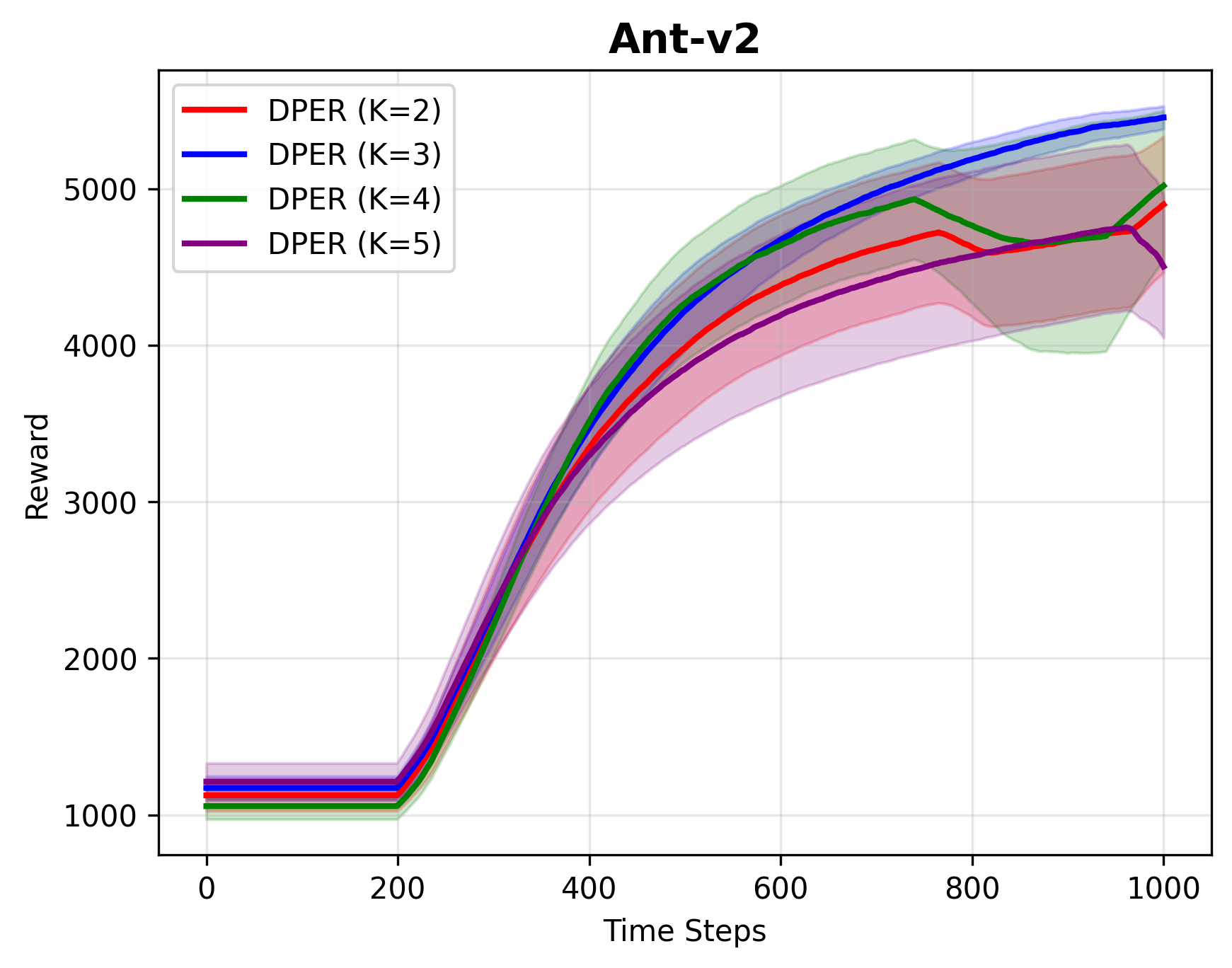}%
    \label{fig:k_reward_ant}
  }\hfill
  \subfigure[BipedalWalker-v3]{%
    \includegraphics[width=0.32\textwidth]{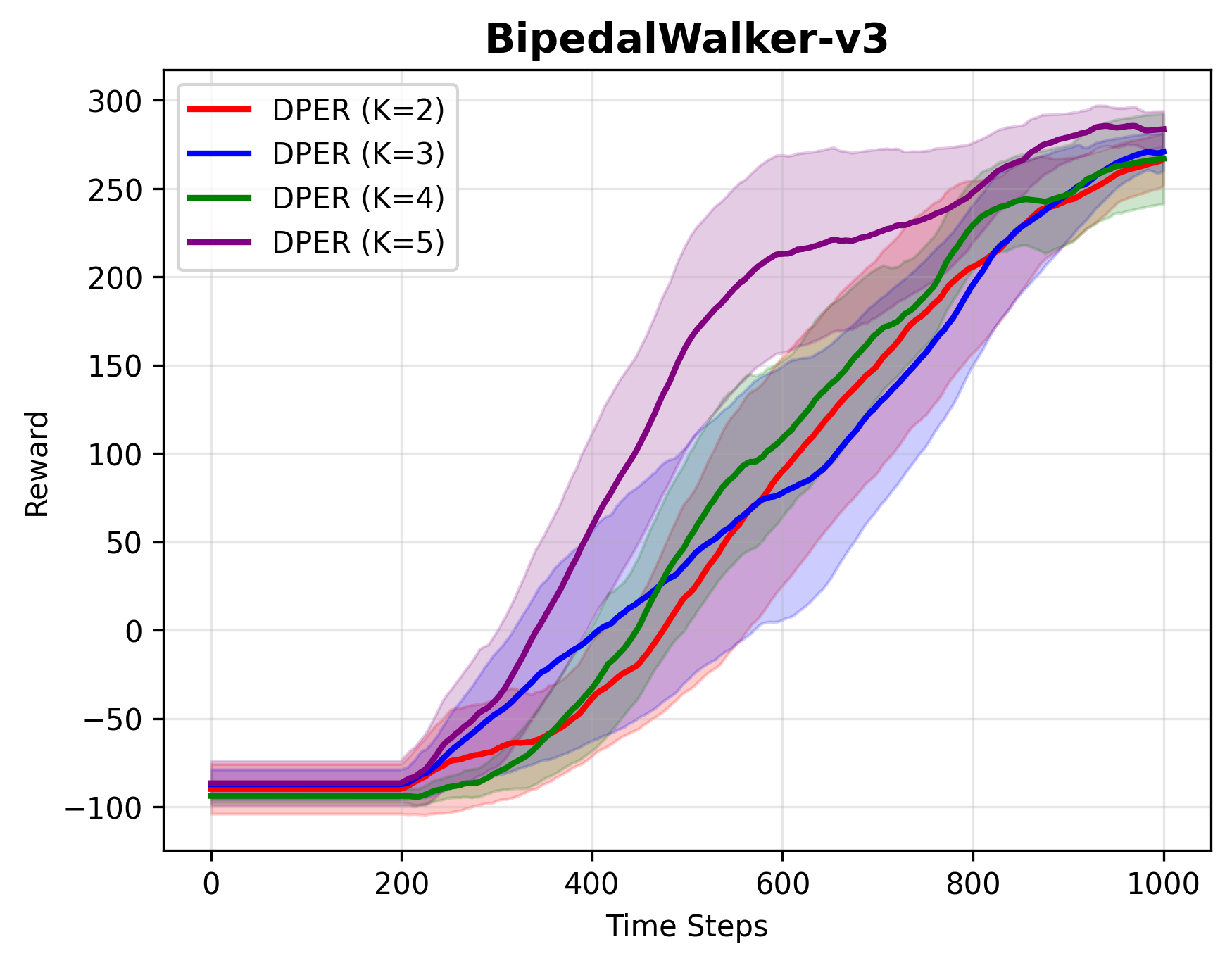}%
    \label{fig:k_reward_bw}
  }\hfill
  \subfigure[HalfCheetah-v2]{%
    \includegraphics[width=0.32\textwidth]{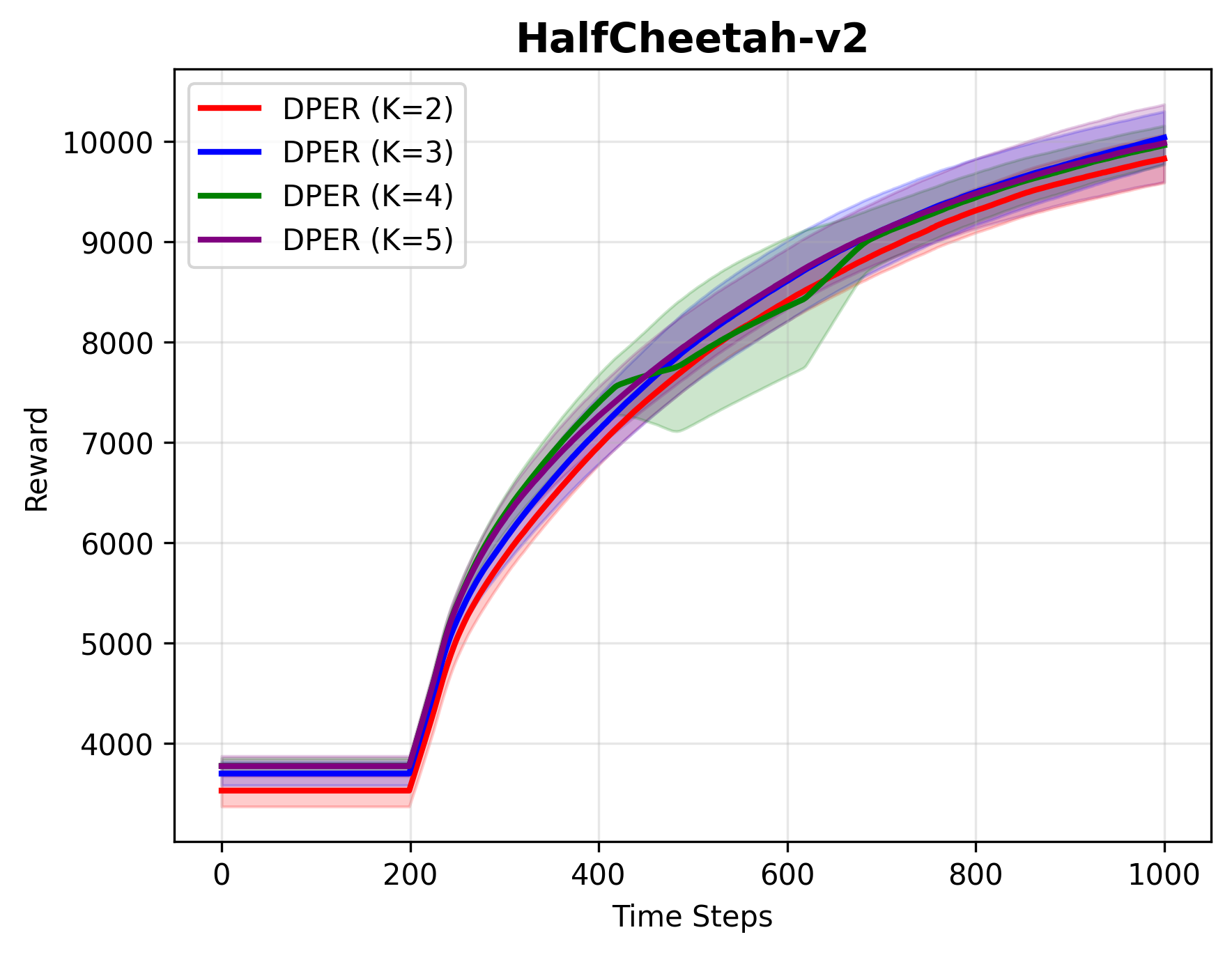}%
    \label{fig:k_reward_hc}
  }

  \vspace{0.6em}

  \subfigure[Hopper-v2]{%
    \includegraphics[width=0.32\textwidth]{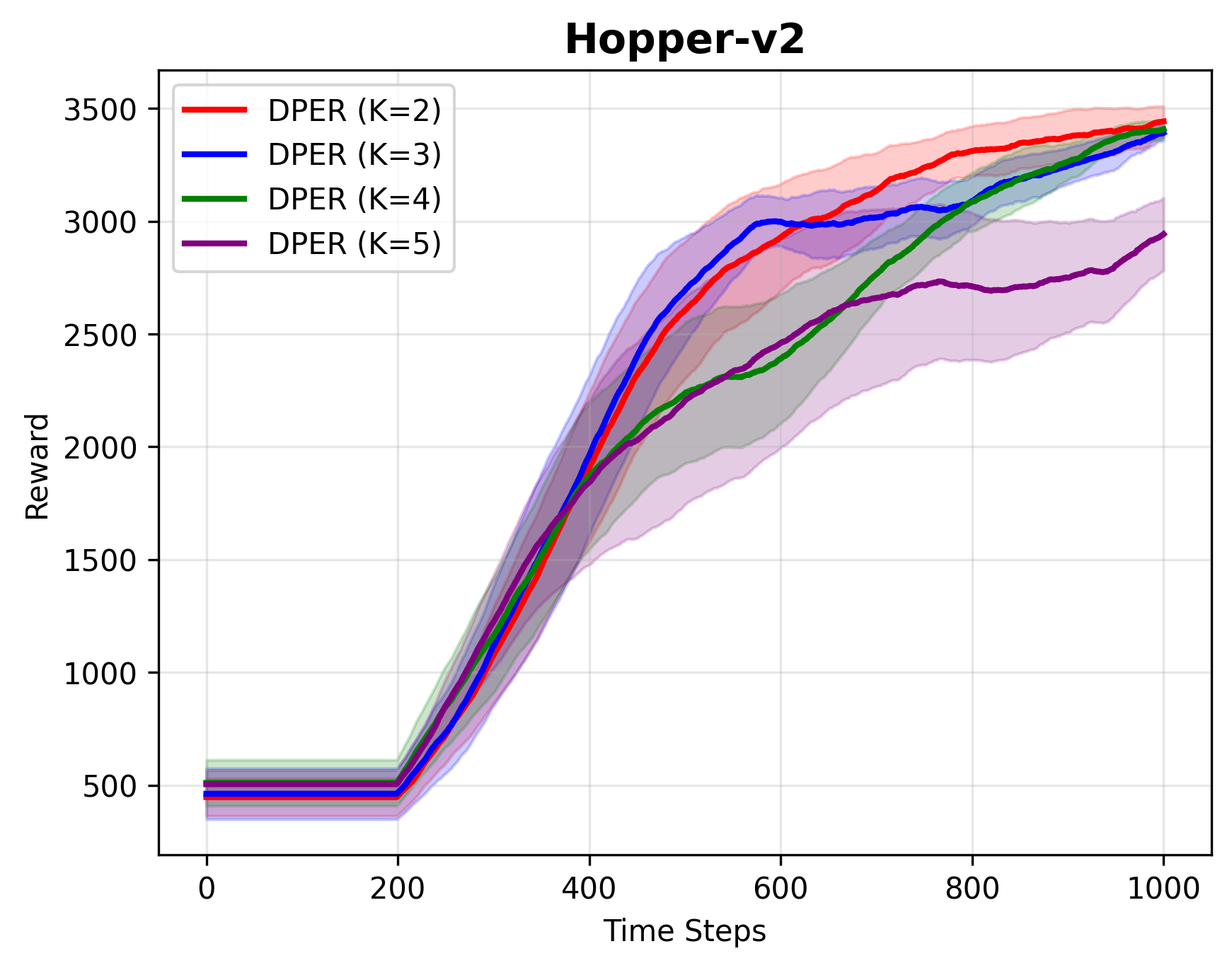}%
    \label{fig:k_reward_hopper}
  }\hfill
  \subfigure[LunarLanderContinuous-v2]{%
    \includegraphics[width=0.32\textwidth]{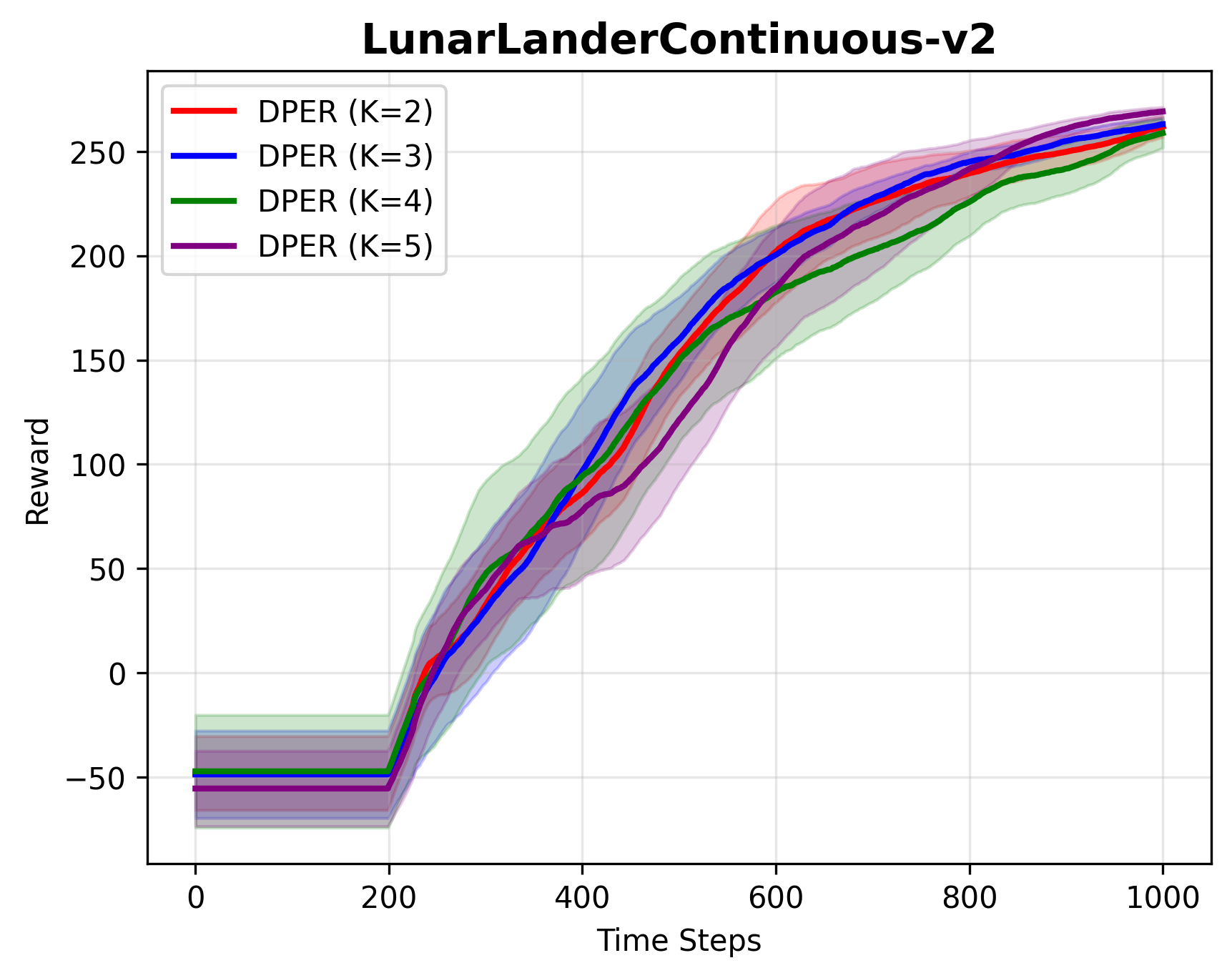}%
    \label{fig:k_reward_llc}
  }\hfill
  \subfigure[Walker2d-v2]{%
    \includegraphics[width=0.32\textwidth]{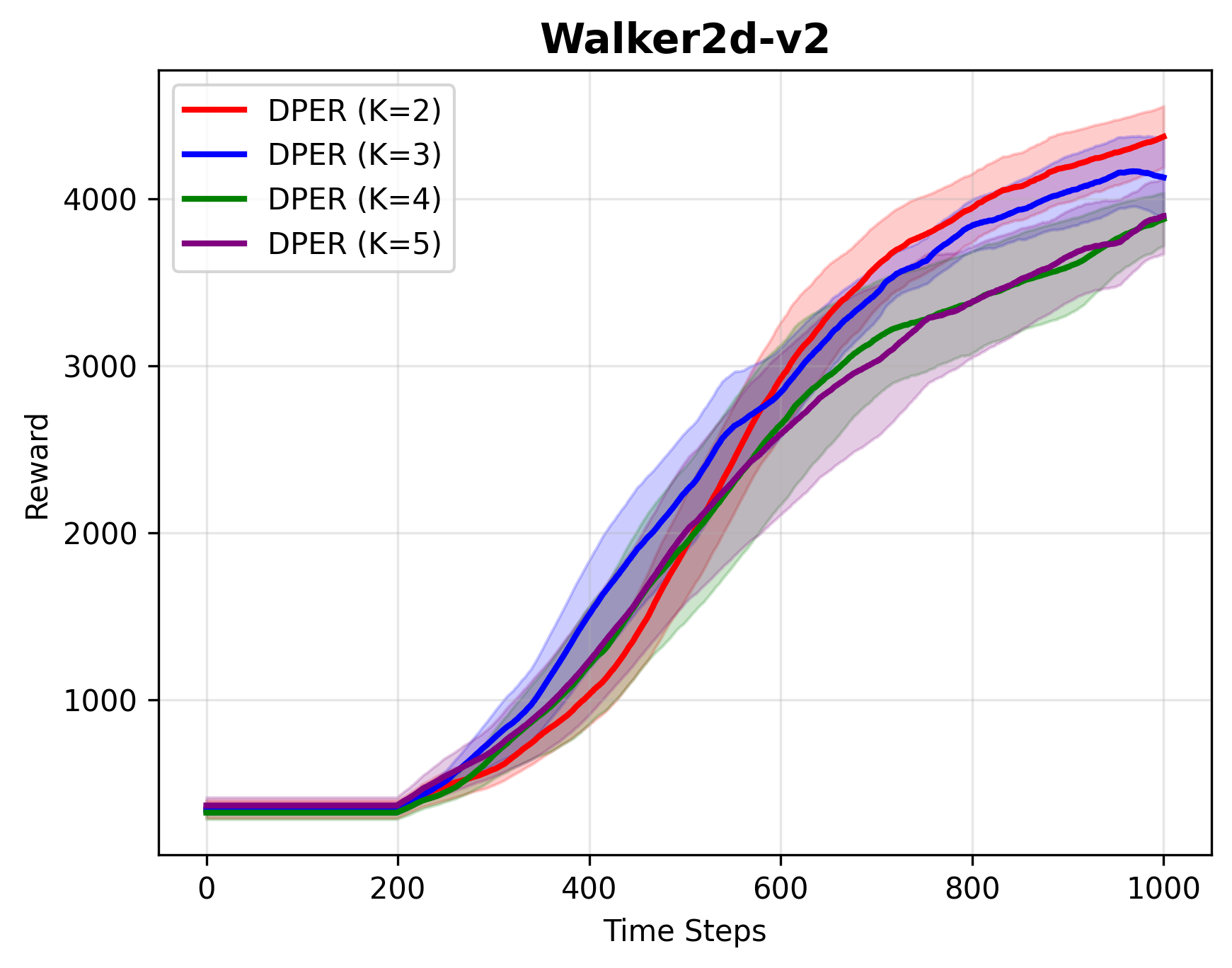}%
    \label{fig:k_reward_walker}
  }

  \caption{K-sweep ablation: average cumulative rewards across tasks; shaded regions indicate half standard deviation.}
  \label{fig:k_rewards_all}
\end{figure}

We further analyze the sensitivity of DPER to the hyperparameter $K$, which dictates the number of candidate batches sampled to find a near on-policy batch for the actor update. Fig. \ref{fig:k_rewards_all} shows the performance of DPER for $K$ values ranging from 2 to 5.

Across all tested environments, we observe no clear, monotonic correlation between the value of $K$ and the final performance of the agent. This counter-intuitive result can be attributed to two main factors. First, our method for identifying on-policy batches relies on approximations, specifically the modeling of the Transition Generator and the subsequent KL divergence calculation. Second, even our largest tested value, $K=5$, represents a tiny fraction of the one million transitions stored in the replay buffer. Therefore, searching within this small window may not be extensive enough to reveal a consistent trend.

However, this finding has a significant and positive practical implication: the performance benefits of DPER do not require a computationally expensive search over the replay buffer. A small value, i.e., $K=2$ or $K=3$, is sufficient to achieve competitive or superior performance across all tasks. This indicates that our batch selection heuristic is effective at quickly identifying a good enough batch that is sufficiently close to the on-policy distribution to ensure stable actor updates. This makes DPER effective while also not being computationally expensive, presenting an advantage for practical applications where high computational loads are undesirable.

\section{Critic Sampling Ablation}
\label{app:critic-sampler}

\begin{figure}[t]
  \centering

  \subfigure[Ant-v2]{%
    \includegraphics[width=0.32\textwidth]{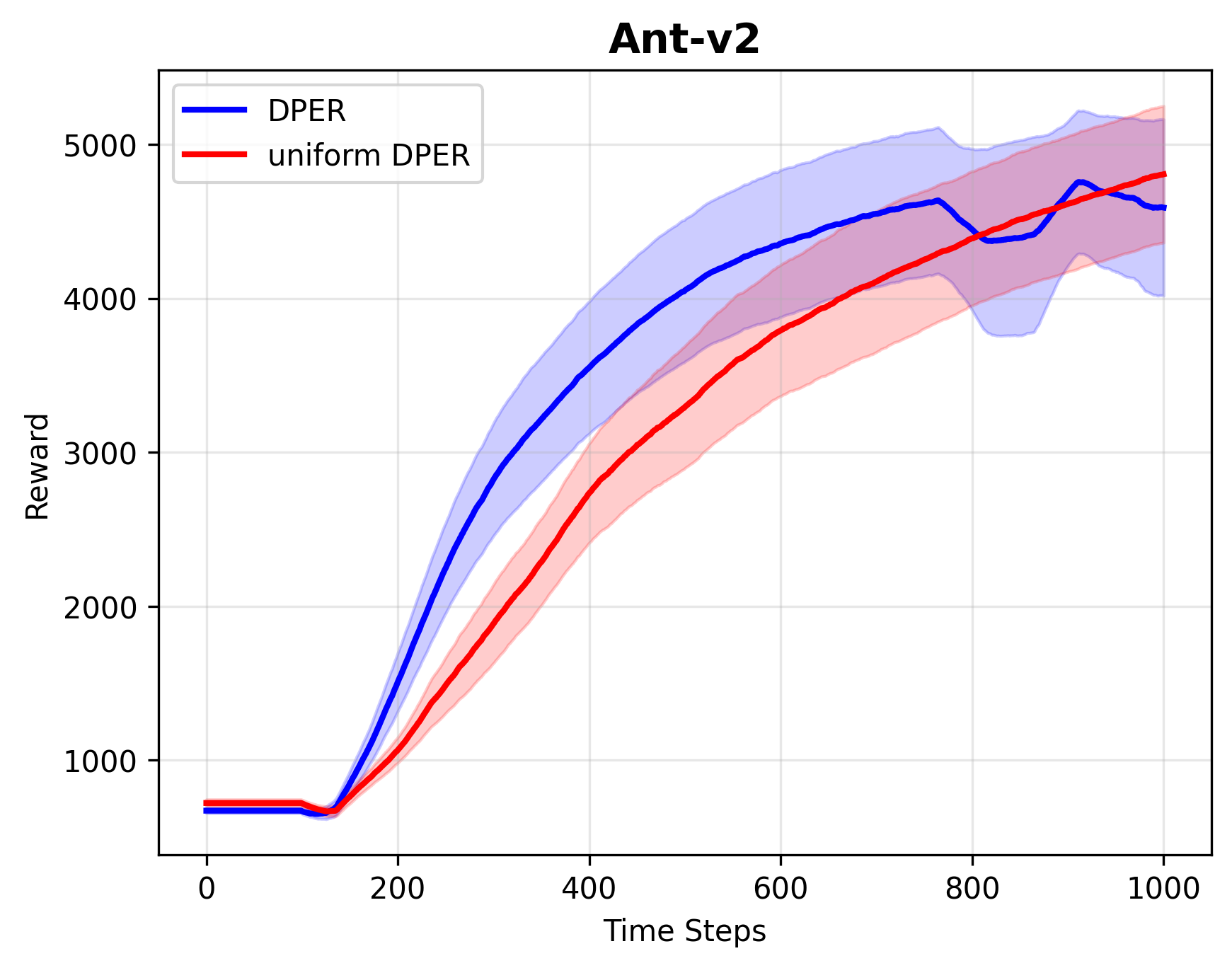}%
    \label{fig:uni_reward_ant}
  }\hfill
  \subfigure[BipedalWalker-v3]{%
    \includegraphics[width=0.32\textwidth]{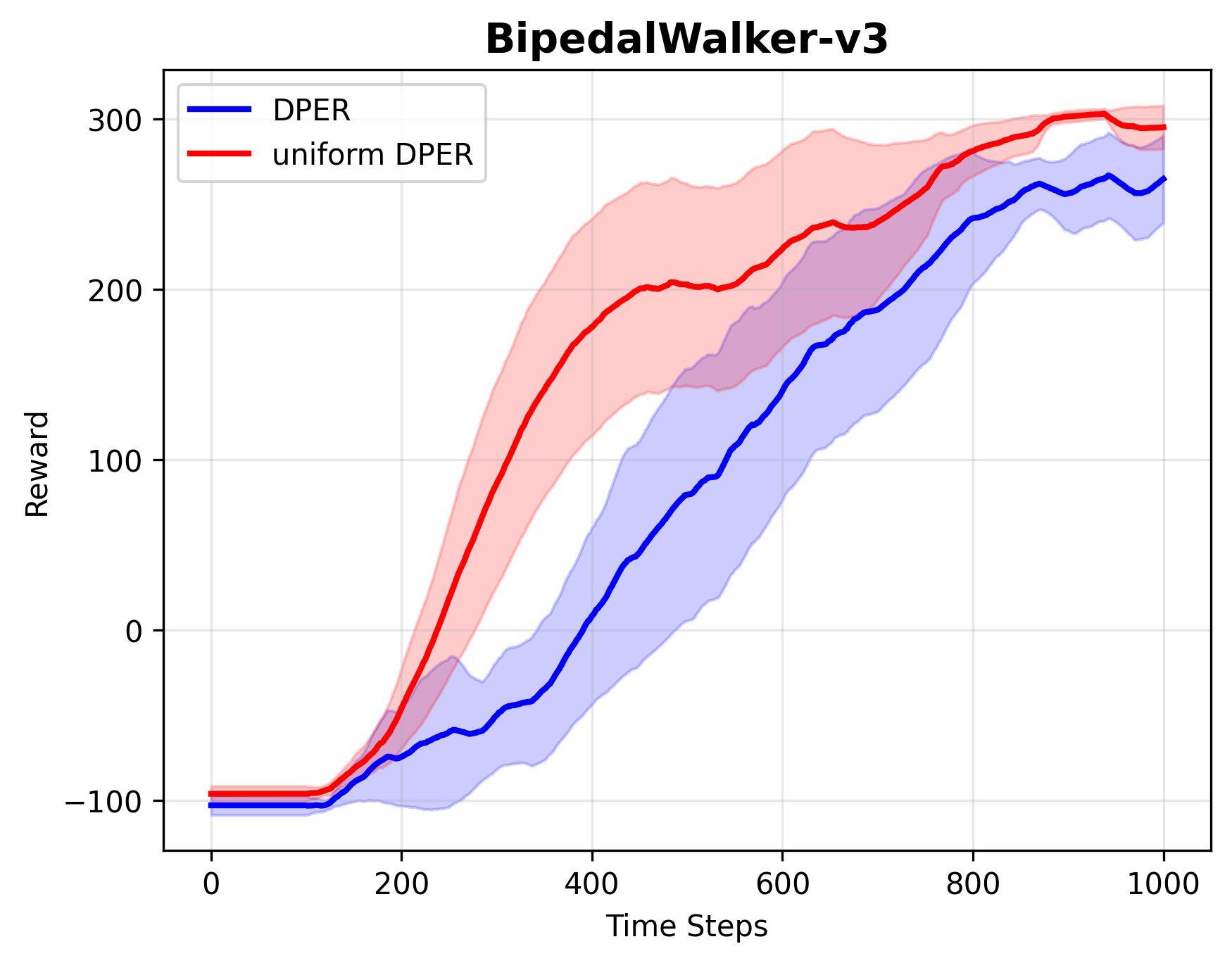}%
    \label{fig:uni_reward_bw}
  }\hfill
  \subfigure[HalfCheetah-v2]{%
    \includegraphics[width=0.32\textwidth]{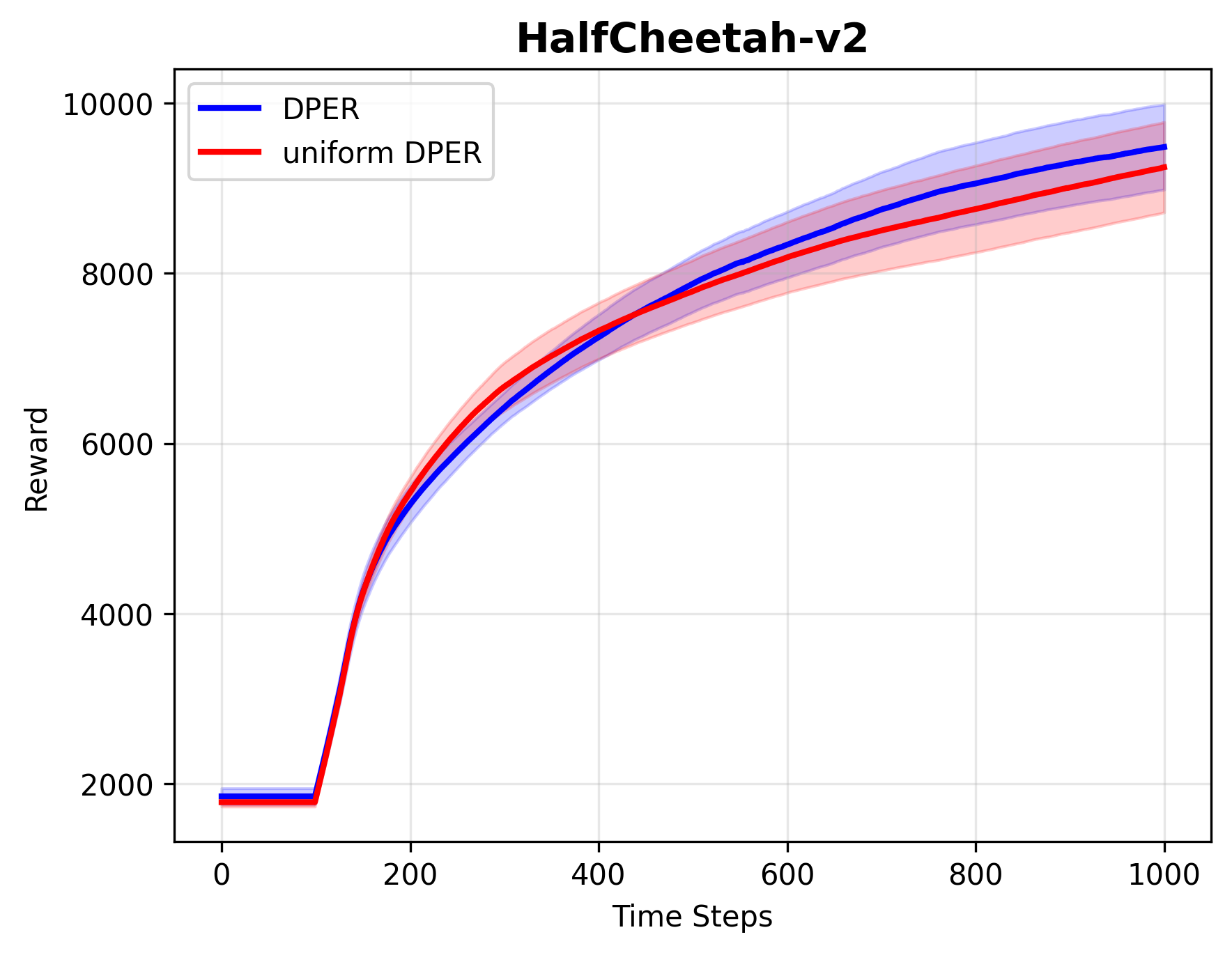}%
    \label{fig:uni_reward_hc}
  }

  \vspace{0.6em}

  \subfigure[Hopper-v2]{%
    \includegraphics[width=0.32\textwidth]{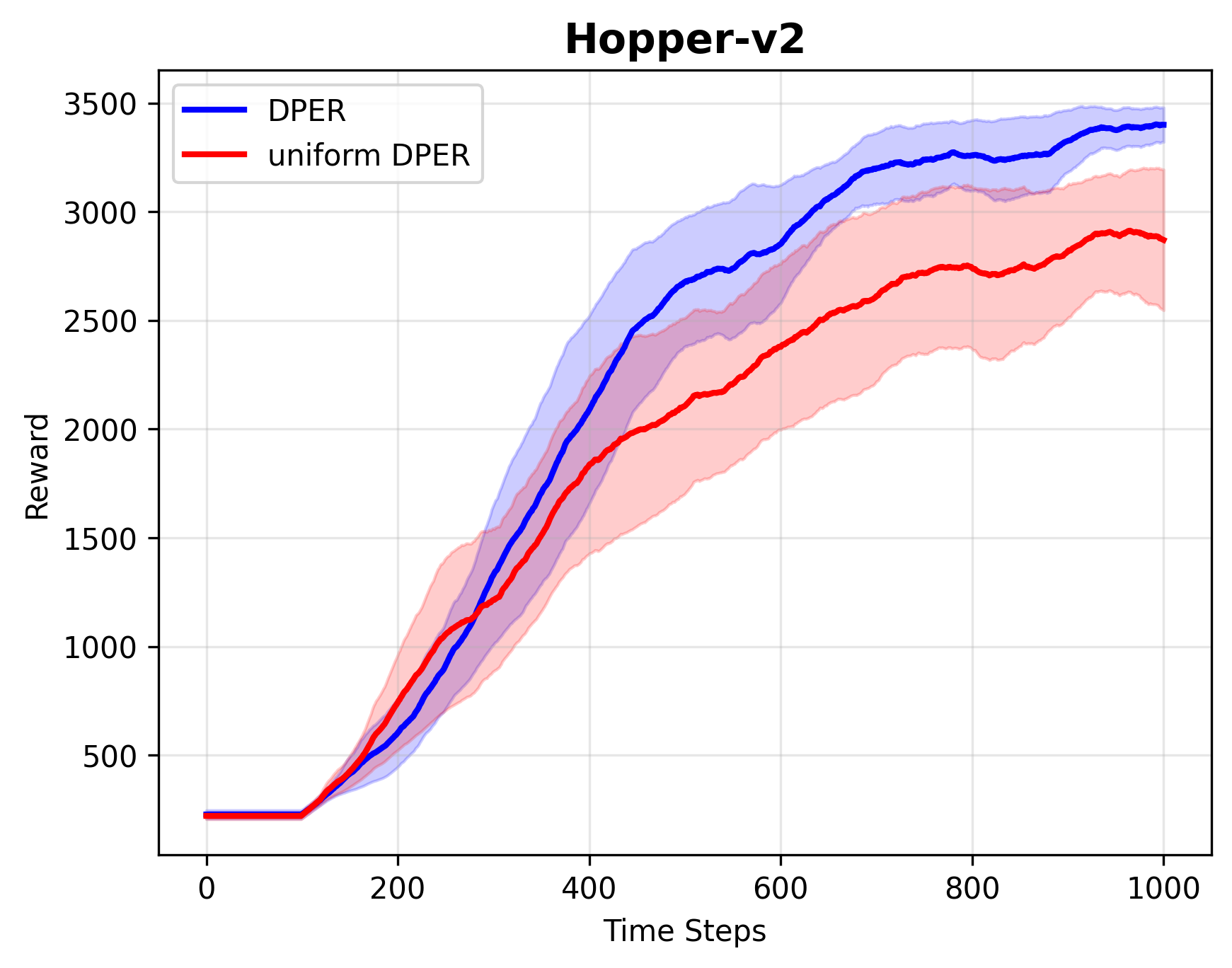}%
    \label{fig:uni_reward_hopper}
  }\hfill
  \subfigure[LunarLanderContinuous-v2]{%
    \includegraphics[width=0.32\textwidth]{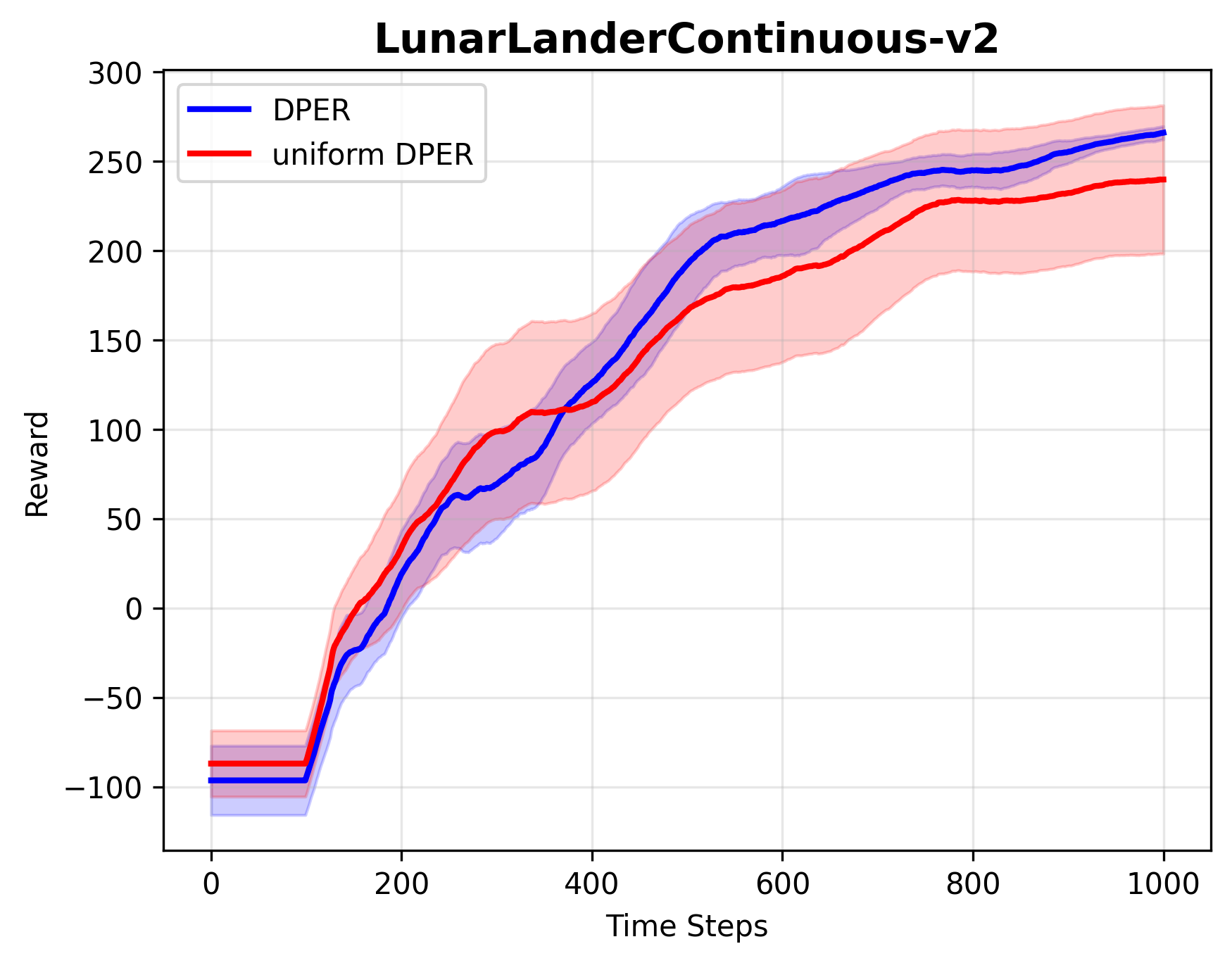}%
    \label{fig:uni_reward_llc}
  }\hfill
  \subfigure[Walker2d-v2]{%
    \includegraphics[width=0.32\textwidth]{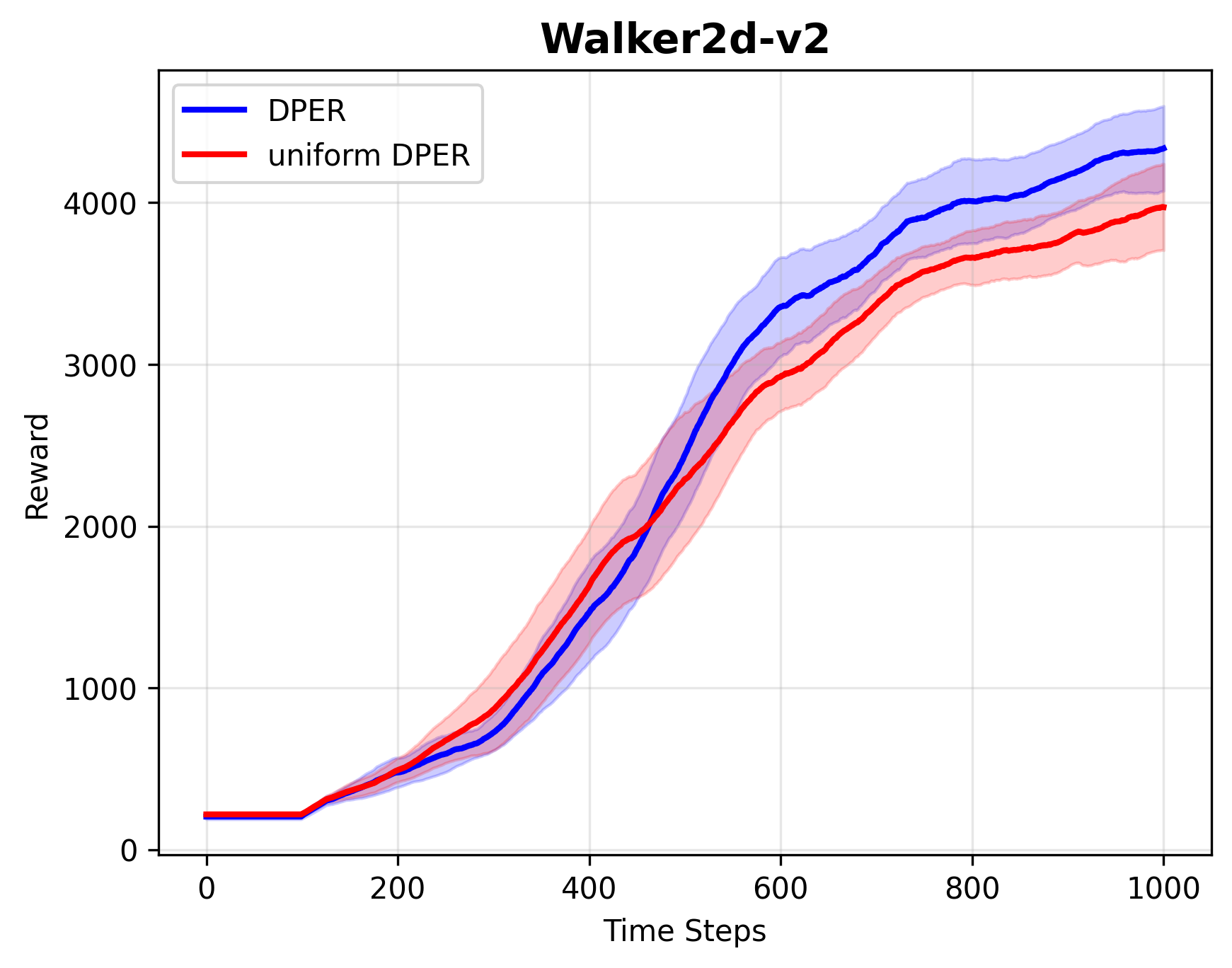}%
    \label{fig:uni_reward_walker}
  }

  \caption{Uniform-DPER: average cumulative rewards across tasks; shaded regions indicate half standard deviation.}
  \label{fig:uni_rewards_all}
\end{figure}

An important insight of our work is that decoupling the experience replay mechanisms for the actor and critic can enhance learning performance. While our DPER algorithm employs prioritized sampling for the critic to accelerate learning, our framework is flexible. To investigate the impact of this choice, we conduct an ablation study comparing our standard DPER approach against a variant that uses uniform random sampling for the critic, which we term uniform DPER. For a fair and computationally efficient comparison, both variants use a fixed hyperparameter of $K=2$ for the actor's batch selection. The results of these experiments are presented in Fig. \ref{fig:uni_rewards_all}.

An important observation from these experiments is that both DPER variants successfully learn across all environments. This contrasts the performance of standard PER, which, as shown in Figures \ref{fig:reward_bw} and \ref{fig:reward_llc}, experienced a learning failure in the BipedalWalker and LunarLanderContinuous tasks. By decoupling the actor's updates from the critic's samples, DPER prevents instability, irrespective of whether the critic itself uses uniform or prioritized sampling. This provides strong evidence that the decoupled sampling strategy is an important factor in robustness of our method.

We observe that both variants of DPER are superior to the other in certain tasks. This observation aligns with our earlier findings where neither vanilla ER nor PER was universally superior across the six environments we used. In summary, this ablation study validates that decoupling is key to stable learning and that the choice of critic sampling within the DPER framework serves as a valuable, task-specific tuning parameter rather than a fixed choice.

\section{Runtime analysis}
\label{app:runtime}

To evaluate the practical viability of DPER, we analyze its computational overhead in comparison to the ER and PER baselines. We measured the total wall-clock time for each algorithm to complete one million training steps, with the results averaged over runs. The mean and standard deviation of these measurements are presented in Table \ref{tab:training_times}.

As expected, vanilla ER is the most computationally efficient method due to its simple random sampling mechanism. The introduction of prioritization in PER incurs a notable overhead, approximately doubling the training time compared to ER. This additional cost stems from the computation of TD errors for all transitions in the replay buffer and the maintenance of the priority data structure.

The computational cost of DPER, as detailed in Algorithm \ref{alg:DPER}, is primarily influenced by the hyperparameter K, the number of candidate batches. The core of DPER's additional workload lies in the loop from lines 8 to 15, where for each of the K candidate batches, the algorithm must:
\begin{enumerate}
    \item Sample a mini-batch $D_n$
    \item Perform a forward pass through the actor network to compute current actions  $\hat{\bm{X}}_n$
    \item Calculate the sample mean and covariance of the action differences
    \item Compute the KL divergence $\eta_n$.
\end{enumerate}

This sequence of operations is performed K times for every single actor update. Consequently, the total training time for DPER exhibits a clear linear relationship with K. As shown in Table \ref{tab:training_times}, each increment in K adds approximately 4,000 seconds to the total training time.

Crucially, even with the smallest tested value, K=2, DPER is computationally more expensive than PER. This is an important trade-off to consider. However, our hyperparameter sensitivity analysis \ref{app:k} demonstrates that DPER achieves superior or competitive performance with a small $K$ (e.g., 2 or 3). This is significant as, while DPER introduces a computational overhead, it does not require an exhaustive search over many candidate batches to be effective. The performance gains documented in our comparative analysis are therefore achievable at a reasonable computational cost.

\begin{table}[h!]
\centering
\begin{tabular}{|l|r|r|}
\hline
\textbf{Algorithm} & \textbf{Mean time (s)} & \textbf{Std time (s)} \\
\hline
DPER K=2 & 18664.84 & 590.51 \\
DPER K=3 & 22549.39 & 604.75 \\
DPER K=4 & 26557.94 & 768.79 \\
DPER K=5 & 30318.89 & 644.03 \\
ER       &  6112.32 & 489.43 \\
PER      & 13924.88 & 995.57 \\
\hline
\end{tabular}
\caption{Mean and standard deviation of training times for different algorithms.}
\label{tab:training_times}
\end{table}

\end{appendices}
\bibliography{sn-article}

\end{document}